\documentclass[runningheads]{llncs}
\usepackage{graphicx}
\usepackage{amsmath,amssymb} 
\usepackage{color}
\usepackage{rotating}
\usepackage[font=small]{caption}

\begin{document}
\pagestyle{headings}
\mainmatter

\title{
   Reconstructive Sparse Code Transfer for
   Contour Detection and Semantic Labeling
}
\titlerunning{Reconstructive Sparse Code Transfer}
\authorrunning{Maire, Yu, Perona}
\author{
   Michael Maire$^{1,2}$, Stella X. Yu$^3$, and Pietro Perona$^2$\\
}
\institute{
   $^1$TTI Chicago\quad
   $^2$California Institute of Technology\\
   $^3$University of California at Berkeley / ICSI\\
   {\tt\small mmaire@ttic.edu},
   {\tt\small stellayu@berkeley.edu},
   {\tt\small perona@caltech.edu}
}
\maketitle

\begin{abstract}
   We frame the task of predicting a semantic labeling as a sparse
   reconstruction procedure that applies a target-specific learned transfer
   function to a generic deep sparse code representation of an image.  This
   strategy partitions training into two distinct stages.  First, in an
   unsupervised manner, we learn a set of dictionaries optimized for sparse
   coding of image patches.  These generic dictionaries minimize error with
   respect to representing image appearance and are independent of any
   particular target task.  We train a multilayer representation via
   recursive sparse dictionary learning on pooled codes output by earlier
   layers.  Second, we encode all training images with the generic dictionaries
   and learn a transfer function that optimizes reconstruction of patches
   extracted from annotated ground-truth given the sparse codes of their
   corresponding image patches.  At test time, we encode a novel image using
   the generic dictionaries and then reconstruct using the transfer function.
   The output reconstruction is a semantic labeling of the test image.
   
   \setlength\parindent{12pt}
   Applying this strategy to the task of contour detection, we demonstrate
   performance competitive with state-of-the-art systems.  Unlike almost all
   prior work, our approach obviates the need for any form of hand-designed
   features or filters.  Our model is entirely learned from image and
   ground-truth patches, with only patch sizes, dictionary sizes and
   sparsity levels, and depth of the network as chosen parameters.  To
   illustrate the general applicability of our approach, we also show initial
   results on the task of semantic part labeling of human faces.

   The effectiveness of our data-driven approach opens new avenues for research
   on deep sparse representations.  Our classifiers utilize this representation
   in a novel manner.  Rather than acting on nodes in the deepest layer, they
   attach to nodes along a slice through multiple layers of the network in
   order to make predictions about local patches.  Our flexible combination
   of a generatively learned sparse representation with discriminatively
   trained transfer classifiers extends the notion of sparse reconstruction
   to encompass arbitrary semantic labeling tasks.
\end{abstract}

\section{Introduction}
\label{sec:introduction}

A multitude of recent work establishes the power of learning hierarchical
representations for visual recognition tasks.  Noteworthy examples include
deep autoencoders~\cite{LRMDCCDN:ICML:2012}, deep convolutional networks
\cite{LKF:ISCAS:2010,KSH:NIPS:2012}, deconvolutional networks
\cite{ZTF:ICCV:2011}, hierarchical sparse coding~\cite{YLL:CVPR:2011}, and
multipath sparse coding~\cite{MHMP}.  Though modeling choices and learning
techniques vary, these architectures share the overall strategy of
concatenating coding (or convolution) operations followed by pooling
operations in a repeating series of layers.  Typically, the representation at
the topmost layer (or pooled codes from multiple layers~\cite{MHMP}) serves
as an input feature vector for an auxiliary classifier, such as a support
vector machine (SVM), tasked with assigning a category label to the image.

Our work is motivated by exploration of the information content of the
representation constructed by the rest of the network.  While the topmost or
pooled features robustly encode object category, what semantics can be
extracted from the spatially distributed activations in the earlier network
layers?  Previous work attacks this question through development of tools for
visualizing and probing network behavior~\cite{ZF:ECCV:2014}.  We
provide a direct result: a multilayer slice above a particular spatial
location contains sufficient information for semantic labeling of a local
patch.  Combining predicted labels across overlapping patches yields a
semantic segmentation of the entire image.

In the case of contour detection (regarded as a binary labeling problem), we
show that a single layer sparse representation (albeit over multiple image
scales and patch sizes) suffices to recover most edges, while a second layer
adds the ability to differentiate (and suppress) texture edges.  This suggests
that contour detection (and its dual problem, image segmentation
\cite{gPb-UCM}) emerge implicitly as byproducts of deep representations.

Moreover, our reconstruction algorithm is not specific to contours.  It is a
recipe for transforming a generic sparse representation into a task-specific
semantic labeling.  We are able to reuse the same multilayer network structure
for contours in order to train a system for semantic segmentation of human
faces.

We make these claims in the specific context of the multipath sparse coding
architecture of Bo~\emph{et al.}~\cite{MHMP}.  We learn sparse codes for different
patch resolutions on image input, and, for deeper layers, on pooled and
subsampled sparse representations of earlier layers.  However, instead of a
final step that pools codes into a single feature vector for the entire
image, we use the distributed encoding in the setting of sparse reconstruction.
This encoding associates a high-dimensional sparse feature vector with each
pixel.  For the traditional image denoising reconstruction task, convolving
these vectors with the patch dictionary from the encoding stage and averaging
overlapping areas yields a denoised version of the original
image~\cite{EA:TIP:2006}.

\begin{figure*}
   \begin{center}
      \begin{minipage}[t]{0.025\linewidth}
         \vspace{3.2\linewidth}
         \begin{sideways}\footnotesize{\textbf{\textsf{Image}}}\end{sideways}
      \end{minipage}
      \begin{minipage}[t]{0.32\linewidth}
         \vspace{0.02\linewidth}
         \setlength\fboxsep{0pt}
         \fbox{\includegraphics[width=1.00\linewidth]{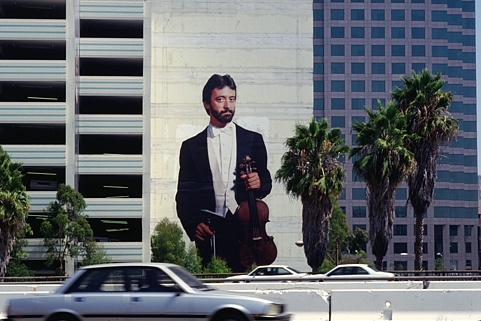}}
      \end{minipage}
      \hspace{0.01\linewidth}
      \begin{minipage}[t]{0.050\linewidth}
         \vspace{2.1\linewidth}
         \begin{center}\LARGE{\textbf{\textsf{$\Rightarrow$}}}\end{center}
      \end{minipage}
      \hspace{-0.01\linewidth}
      \begin{minipage}[t]{0.23\linewidth}
         \vspace{0pt}
         \begin{center}
            \footnotesize{\textbf{\textsf{Patch Dictionary}}}\\
            \includegraphics[width=0.75\linewidth]{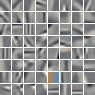}\\
            \footnotesize{\textbf{\textsf{Batch OMP}}}
         \end{center}
      \end{minipage}
      \hspace{-0.01\linewidth}
      \begin{minipage}[t]{0.050\linewidth}
         \vspace{2.1\linewidth}
         \begin{center}\LARGE{\textbf{\textsf{$\Rightarrow$}}}\end{center}
      \end{minipage}
      \hspace{0.005\linewidth}
      \begin{minipage}[t]{0.22\linewidth}
         \vspace{0.12\linewidth}
         \begin{center}
            \begin{sideways}\footnotesize{\textit{\textsf{~coefficients $\rightarrow$}}}\end{sideways}
            {\vrule width\fboxrule\vbox{\hbox{\includegraphics[width=0.75\linewidth]{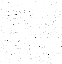}}\hrule height\fboxrule}}\\
            \footnotesize{\textit{\textsf{\quad pixels $\rightarrow$}}}
         \end{center}
      \end{minipage}
      \begin{minipage}[t]{0.025\linewidth}
         \vspace{1.20\linewidth}
         \begin{rotate}{270}\footnotesize{\textbf{\textsf{Sparse Codes}}}\end{rotate}
      \end{minipage} \\
      \vspace{0.03\linewidth}
      \begin{minipage}[t]{1.0\linewidth}
         \vspace{0pt}
         \begin{center}\rule{1.0\linewidth}{1.0pt}\end{center}
      \end{minipage} \\
      \vspace{0.02\linewidth}
      \begin{minipage}[t]{0.025\linewidth}
         \vspace{1.50\linewidth}
         \begin{sideways}\footnotesize{\textbf{\textsf{Sparse Codes}}}\end{sideways}
      \end{minipage}
      \begin{minipage}[t]{0.22\linewidth}
         \vspace{0.12\linewidth}
         \begin{center}
            \begin{sideways}\footnotesize{\textit{\textsf{~coefficients $\rightarrow$}}}\end{sideways}
            {\vrule width\fboxrule\vbox{\hbox{\includegraphics[width=0.75\linewidth]{figs/overview-g_mx_abs.png}}\hrule height\fboxrule}}\\
            \footnotesize{\textit{\textsf{\quad pixels $\rightarrow$}}}
         \end{center}
      \end{minipage}
      \hspace{0.005\linewidth}
      \begin{minipage}[t]{0.050\linewidth}
         \vspace{2.1\linewidth}
         \begin{center}\LARGE{\textbf{\textsf{$\Rightarrow$}}}\end{center}
      \end{minipage}
      \hspace{-0.01\linewidth}
      \begin{minipage}[t]{0.23\linewidth}
         \vspace{0pt}
         \begin{center}
            \footnotesize{\textbf{\textsf{Patch Dictionary}}}\\
            \includegraphics[width=0.75\linewidth]{figs/overview-D.png}\\
            \footnotesize{\textbf{\textsf{Reconstruction}}}
         \end{center}
      \end{minipage}
      \hspace{-0.015\linewidth}
      \begin{minipage}[t]{0.050\linewidth}
         \vspace{2.1\linewidth}
         \begin{center}\LARGE{\textbf{\textsf{$\Rightarrow$}}}\end{center}
      \end{minipage}
      \hspace{0.005\linewidth}
      \begin{minipage}[t]{0.32\linewidth}
         \vspace{0.02\linewidth}
         \setlength\fboxsep{0pt}
         \fbox{\includegraphics[width=1.00\linewidth]{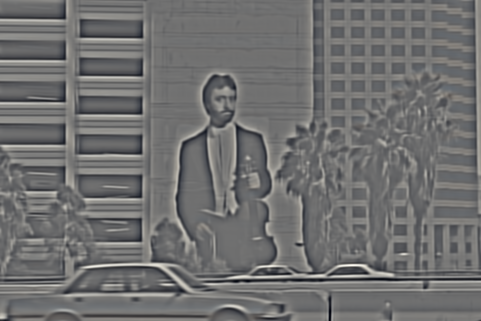}}
         \vspace{-0.15\linewidth}
         \begin{center}\footnotesize{\textbf{\textsf{Reconstructed Image\\(locally zero-mean)}}}\end{center}
         \vspace{-0.15\linewidth}
      \end{minipage} \\
      \vspace{0.03\linewidth}
      \begin{minipage}[t]{1.0\linewidth}
         \vspace{0pt}
         \hspace{0.14\linewidth}
         \begin{sideways}\LARGE{\textbf{\textsf{$\Leftarrow$}}}\end{sideways}
      \end{minipage} \\
      \vspace{0.02\linewidth}
      \begin{minipage}[t]{0.025\linewidth}
         \vspace{2.50\linewidth}
         \begin{sideways}\footnotesize{\textbf{\textsf{Rectified Sparse Codes}}}\end{sideways}
      \end{minipage}
      \begin{minipage}[t]{0.23\linewidth}
         \vspace{0.12\linewidth}
         \begin{center}
            \begin{sideways}\hspace{0.40\linewidth}\footnotesize{\textit{\textsf{~coefficients $\rightarrow$}}}\end{sideways}
            {\vrule width\fboxrule\vbox{\hbox{\includegraphics[width=0.75\linewidth]{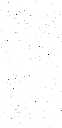}}\hrule height\fboxrule}}\\
            \footnotesize{\textit{\textsf{\quad pixels $\rightarrow$}}}
         \end{center}
      \end{minipage}
      \hspace{0.005\linewidth}
      \begin{minipage}[t]{0.050\linewidth}
         \vspace{3.9\linewidth}
         \begin{center}\LARGE{\textbf{\textsf{$\Rightarrow$}}}\end{center}
      \end{minipage}
      \hspace{-0.015\linewidth}
      \begin{minipage}[t]{0.23\linewidth}
         \vspace{0pt}
         \begin{center}
            \footnotesize{\textbf{\textsf{Transfer Dictionary}}}\\
            \includegraphics[width=0.75\linewidth]{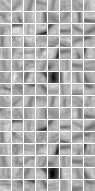}\\
            \footnotesize{\textbf{\textsf{Reconstruction}}}
         \end{center}
      \end{minipage}
      \hspace{-0.015\linewidth}
      \begin{minipage}[t]{0.050\linewidth}
         \vspace{3.9\linewidth}
         \begin{center}\LARGE{\textbf{\textsf{$\Rightarrow$}}}\end{center}
      \end{minipage}
      \hspace{0.005\linewidth}
      \begin{minipage}[t]{0.32\linewidth}
         \vspace{0.20\linewidth}
         \setlength\fboxsep{0pt}
         \fbox{\includegraphics[width=1.00\linewidth]{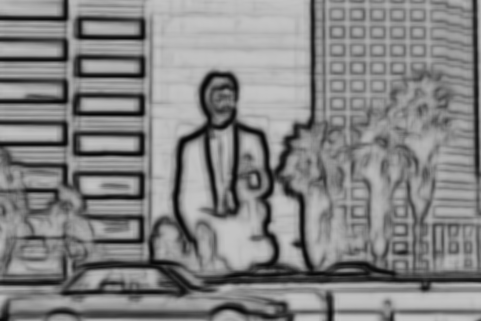}}
         \vspace{-0.15\linewidth}
         \begin{center}\footnotesize{\textbf{\textsf{Contour Detection}}}\end{center}
      \end{minipage}
   \end{center}
   \caption{
      \textbf{Reconstructive sparse code transfer.}
      \emph{Top:}
         Applying batch orthogonal matching pursuit (BOMP)~\cite{OMP,BOMP}
         against a learned appearance dictionary determines a sparse code
         for each patch in the image.  We subtract means of patch RGB
         channels prior to encoding.
      \emph{Bottom:}
         Convolving the sparse code representation with the same dictionary
         reconstructs a locally zero-mean version of the input image.
         Alternatively, rectifying the representation and applying a transfer
         function (learned for contour detection) reconstructs an edge map.
         For illustrative purposes, we show a small single-layer dictionary
         (64 11x11 patches) and simple transfer functions (logistic
         classifiers) whose coefficients are rendered as a corresponding
         dictionary.  Our deeper representations (Figure~\ref{fig:network})
         are much higher-dimensional and yield better performing, but not
         easily visualized, transfer functions.
   }
   \label{fig:overview}
\end{figure*}

\begin{figure*}[t!]
   \begin{center}
      \begin{minipage}[t]{0.20\linewidth}
         \vspace{0pt}
         \setlength\fboxsep{0pt}
         \begin{center}
            \footnotesize{\textbf{\textsf{Multiple Scales}}}
            \vspace{0.015\linewidth}\\
            \fbox{\includegraphics[width=0.32\linewidth]{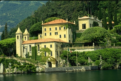}}\hfill
            \fbox{\includegraphics[width=0.65\linewidth]{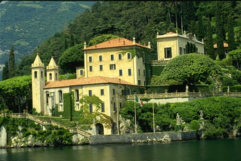}}\\
            \fbox{\includegraphics[width=1.00\linewidth]{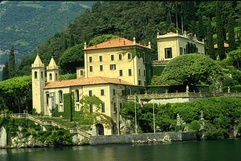}}
         \end{center}
      \end{minipage}
      \begin{minipage}[t]{0.050\linewidth}
         \vspace{2.5\linewidth}
         \begin{center}\LARGE{\textbf{\textsf{$\times$}}}\end{center}
      \end{minipage}
      \hspace{-0.025\linewidth}
      \begin{minipage}[t]{0.30\linewidth}
         \vspace{0pt}
         \setlength\fboxsep{1pt}
         \begin{center}
            \footnotesize{\textbf{\textsf{Layer 1 Dictionaries}}}\\
            \vspace{0.01\linewidth}
            \fbox{%
               \setlength\fboxsep{2pt}%
               $\begin{array}{c}%
                  \fbox{\scriptsize{\textsf{5x5 patches, 64 atoms}}}%
                  \vspace{0.01\linewidth}\\%
                  \fbox{\scriptsize{\textsf{11x11 patches, 64 atoms}}}%
               \end{array}$%
            }\\
            \fbox{%
               \setlength\fboxsep{2pt}%
               $\begin{array}{c}%
                  \fbox{\scriptsize{\textsf{5x5 patches, 512 atoms}}}%
                  \vspace{0.01\linewidth}\\%
                  \fbox{\scriptsize{\textsf{11x11 patches, 512 atoms}}}%
                  \vspace{0.01\linewidth}\\%
                  \fbox{\scriptsize{\textsf{21x21 patches, 512 atoms}}}%
                  \vspace{0.01\linewidth}\\%
                  \fbox{\scriptsize{\textsf{31x31 patches, 512 atoms}}}%
               \end{array}$%
            }
         \end{center}
      \end{minipage}
      \hspace{-0.03\linewidth}
      \begin{minipage}[t]{0.10\linewidth}
         \vspace{0pt}
         \begin{center}
            \vspace{0.19\linewidth}
            \scriptsize{\textit{\textsf{pooling}}}\\
            \vspace{-0.25\linewidth}
            \Large{\textbf{\textsf{$\longrightarrow$}}}\\
            \vspace{-0.15\linewidth}
            \Large{\textbf{\textsf{$\longrightarrow$}}}\\
            \vspace{0.35\linewidth}
            \LARGE{\textbf{\textsf{$\Longrightarrow$}}}
         \end{center}
      \end{minipage}
      \hspace{-0.03\linewidth}
      \begin{minipage}[t]{0.30\linewidth}
         \vspace{0pt}
         \setlength\fboxsep{1pt}
         \begin{center}
            \footnotesize{\textbf{\textsf{Layer 2 Dictionaries}}}\\
            \vspace{0.01\linewidth}
            \fbox{%
               \setlength\fboxsep{2pt}%
               $\begin{array}{c}%
                  \fbox{\scriptsize{\textsf{5x5 patches, 512 atoms}}}%
                  \vspace{0.01\linewidth}\\%
                  \fbox{\scriptsize{\textsf{5x5 patches, 512 atoms}}}%
               \end{array}$%
            }
         \end{center}
         \begin{center}
            \vspace{-0.10\linewidth}
            \begin{sideways}\LARGE{\textbf{\textsf{$\Leftarrow$}}}\end{sideways}\\
            \vspace{-0.01\linewidth}
            \scriptsize{\textit{\textsf{
               rectify, upsample, concatenate\\
               sparse activation maps
            }}}\\
            \hspace{-0.15\linewidth}\begin{rotate}{240}\LARGE{\textbf{\textsf{$\Rightarrow$}}}\end{rotate}
         \end{center}
      \end{minipage}
   \end{center}
   \begin{center}
      \begin{minipage}[t]{1.00\linewidth}
         \begin{center}
            \includegraphics[width=0.8125\linewidth]{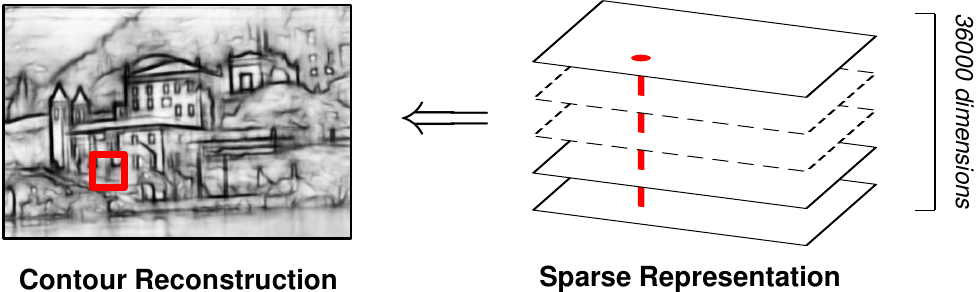}%
         \end{center}
      \end{minipage}
   \end{center}
   \vspace{-0.011\linewidth}
   \caption{
      \textbf{Multipath sparse coding and reconstruction network.}
         We resize an image to $6$ different scales ($3$ shown) and encode
         each using dictionaries learned for different patch sizes
         and atom counts.  Encoding sparsity is $2$ and $4$ nonzero
         coefficients for $64$ and $512$ atom dictionaries, respectively.
         The output representation of the smaller dictionaries is pooled,
         subsampled, and fed to a second sparse coding layer.  We then rectify
         all sparse activation maps, upsample them to the original grid size,
         and concatenate to form a $36000$-dimensional sparse representation
         for each pixel on the image grid.  A set of logistic classifiers
         then transform the sparse vector associated with each pixel (red
         vertical slice) into a predicted labeling of the surrounding patch
         (red box).
   }
   \label{fig:network}
\end{figure*}

Our strategy is to instead swap in an entirely different dictionary for use in
reconstruction.  Here we generalize the notion of ``dictionary'' to include any
function which takes a sparse feature vector as input and outputs predicted
labels for a patch.  Throughout the paper, these transfer dictionaries take the
form of a set of logistic regression functions: one function for predicting
the label of each pixel in the output patch.  For a simplified toy example,
Figure~\ref{fig:overview} illustrates the reconstruction obtained with such a
dictionary learned for the contour detection task.  Figure~\ref{fig:network}
diagrams the much larger multipath sparse coding network that our actual
system uses to generate high-dimensional sparse representations.  The
structural similarity to the multipath network of Bo~\emph{et al.}~\cite{MHMP} is by
design.  They tap part of such a network for object recognition; we tap a
different part of the network for semantic segmentation.  This suggests that
it may be possible to use an underlying shared representation for both tasks.

In addition to being an implicit aspect of deep representations used for object
recognition, our approach to contour detection is entirely free of reliance on
hand-crafted features.  As Section~\ref{sec:related} reviews, this
characteristic is unique amongst competing contour detection algorithms.
Sections~\ref{sec:representation},~\ref{sec:transfer}, and~\ref{sec:multipath}
describe the technical details behind our two-stage approach of sparse coding
and reconstructive transfer.  Section~\ref{sec:experiments} visualizes and
benchmarks results for our primary application of contour detection on the
Berkeley segmentation dataset (BSDS)~\cite{BSDS}.  We also show results for a
secondary application of semantic part labeling on the Labeled Faces in the
Wild (LFW) dataset~\cite{LFW,LFW-parts}.  Section~\ref{sec:conclusion}
concludes.

\section{Related Work}
\label{sec:related}

Contour detection has long been a major research focus in computer vision.
Arbel\'aez \emph{et al.}~\cite{gPb-UCM} catalogue a vast set of historical and
modern algorithms.  Three different approaches~\cite{gPb-UCM,SCG,SE} appear
competitive for state-of-the-art accuracy.  Arbel\'aez \emph{et al.}~\cite{gPb-UCM}
derive pairwise pixel affinities from local color and texture gradients
\cite{Pb} and apply spectral clustering~\cite{NCuts} followed by morphological
operations to obtain a global boundary map.

Ren and Bo~\cite{SCG} adopt the same pipeline, but use gradients of sparse
codes instead of the color and texture gradients developed by Martin
\emph{et al.}~\cite{Pb}.  Note that this is completely different from the manner in
which we propose to use sparse coding for contour detection.  In~\cite{SCG},
sparse codes from a dictionary of small $5 \times 5$ patches serve as
replacement for the textons~\cite{MBLS:IJCV:2001} used in previous
work~\cite{Pb,gPb-UCM}.  Borrowing the hand-designed filtering scheme of
\cite{Pb}, half-discs at multiple orientations act as regions over which
codes are pooled into feature vectors and then classified using an SVM.
In contrast, we use a range of patch resolutions, from $5 \times 5$ to
$31 \times 31$, without hand-designed gradient operations, in a reconstructive
setting through application of a learned transfer dictionary.  Our sparse
codes assume a role different than that of serving as glorified textons.

Doll\'ar and Zitnick~\cite{SE} learn a random decision forest on feature
channels consisting of image color, gradient magnitude at multiple
orientations, and pairwise patch differences.  They cluster ground-truth
edge patches by similarity and train the random forest to predict structured
output.  The emphasis on describing local edge structure in both~\cite{SE}
and previous work~\cite{RFM:ECCV:2006,LZD:CVPR:2013} matches our intuition.
However, sparse coding offers a more flexible methodology for
achieving this goal.  Unlike~\cite{SE}, we learn directly from image
data (not predefined features), in an unsupervised manner, a generic (not
contour-specific) representation, which can then be ported to many tasks via a
second stage of supervised transfer learning.

Mairal \emph{et al.}~\cite{MLBHP:ECCV:2008} use sparse models as the foundation for
developing an edge detector.  However, they focus on discriminative
dictionary training and per-pixel labeling using a linear classifier on
feature vectors derived from error residuals during sparse coding of patches.
This scheme does not benefit from the spatial averaging of overlapping
predictions that occurs in structured output paradigms such as \cite{SE} and
our proposed algorithm.  It also does not incorporate deeper layers of coding,
an aspect we find to be crucial for capturing texture characteristics in the
sparse representation.

Yang \emph{et al.}~\cite{YWLSH:CVPR:2012} study the problem of learning dictionaries
for coupled feature spaces with image super-resolution as an application.  We
share their motivation of utilizing sparse coding in a transfer learning
context.  As the following sections detail, we differ in our choice of a
modular training procedure split into distinct unsupervised (generic)
and supervised (transfer) phases.  We are unique in targeting contour
detection and face part labeling as applications.

\section{Sparse Representation}
\label{sec:representation}

Given image $I$ consisting of $c$ channels ($c=3$ for an RGB color image)
defined over a 2-dimension grid, our sparse coding problem is to represent
each $m \times m \times c$ patch $x \in I$ as a sparse linear combination $z$
of elements from a dictionary
$D = [d_0, d_1, \ldots, d_{L-1}] \in \Re^{(m \cdot m \cdot c) \times L}$.
From a collection of patches $X = [x_0, x_1, ...]$ randomly sampled from a
set of training images, we learn the corresponding sparse representations
$Z = [z_0, z_1, ...]$ as well as the dictionary $D$ using the MI-KSVD algorithm
proposed by Bo \emph{et al.}~\cite{MHMP}.  MI-KSVD finds an approximate solution to
the following optimization problem:
\begin{equation}
   \begin{aligned}
      & \underset{D,~Z}{\mathrm{argmin}}
         \left[
            ||X - DZ||^2_F
            + \lambda \sum_{i=0}^{L-1} \sum_{j=0, j \ne i}^{L-1} |d_i^T d_j|
         \right] \\
      & s.t.~\forall i,~||d_i||_2 = 1~\mathrm{and}~\forall n,~||z_n||_0 \le K
   \end{aligned}
   \label{eq:mi-ksvd}
\end{equation}
where $||\cdot||_F$ denotes Frobenius norm and $K$ is the desired sparsity
level.  MI-KSVD adapts KSVD~\cite{KSVD} by balancing reconstruction error
with mutual incoherence of the dictionary.  This unsupervised training stage
is blind to any task-specific uses of the sparse representation.

Once the dictionary is fixed, the desired encoding $z \in \Re^L$ of a novel
patch $x \in \Re^{m \cdot m \cdot c}$ is:
\begin{equation}
   \underset{z}{\mathrm{argmin}} ||x-Dz||^2 \quad s.t.~||z||_0 \le K
   \label{eq:encoding}
\end{equation}
Obtaining the exact optimal $z$ is NP-hard, but the orthogonal matching
pursuit (OMP) algorithm~\cite{OMP} is a greedy iterative routine that works
well in practice.  Over each of $K$ rounds, it selects the dictionary atom
(codeword) best correlated with the residual after orthogonal projection
onto the span of previously selected codewords.  Batch orthogonal matching
pursuit~\cite{BOMP} precomputes correlations between codewords to significantly
speed the process of coding many signals against the same dictionary.  We
extract the $m \times m$ patch surrounding each pixel in an image and encode
all patches using batch orthogonal matching pursuit.

\section{Dictionary Transfer}
\label{sec:transfer}

Coding an image $I$ as described in the previous section produces a sparse
matrix $Z \in \Re^{L \times N}$, where $N$ is the number of pixels in the image
and each column of $Z$ has at most $K$ nonzeros.  Reshaping each of the $L$
rows of $Z$ into a 2-dimensional grid matching the image size, convolving with
the corresponding codeword from $D$, and summing the results approximately
reconstructs the original image.  Figure~\ref{fig:overview} (middle) shows an
example with the caveat that we drop patch means from the sparse representation
and hence also from the reconstruction.  Equivalently, one can view $D$ as
defining a function that maps a sparse vector $z \in \Re^{L}$ associated with
a pixel to a predicted patch $P \in \Re^{m \times m \times c}$ which is
superimposed on the surrounding image grid and added to overlapping
predictions.

We want to replace $D$ with a function $F(z)$ such that applying this
procedure with $F(\cdot)$ produces overlapping patch predictions that, when
averaged, reconstruct signal $\widehat{G}$ which closely approximates some
desired ground-truth labeling $G$.  $G$ lives on the same 2-dimensional grid
as $I$, but may differ in number of channels.  For contour detection, $G$ is a
single-channel binary image indicating presence or absence of an edge at each
pixel.  For semantic labeling, $G$ may have as many channels as categories
with each channel serving as an indicator function for category presence at
every location.

We regard choice of $F(\cdot)$ as a transfer learning problem given
examples of sparse representations and corresponding ground-truth,
$\{(Z_0,G_0), (Z_1,G_1), \ldots \}$.  To further simplify the problem, we
consider only patch-wise correspondence.  Viewing $Z$ and $G$ as living on the
image grid, we sample a collection of patches $\{g_0, g_1, \ldots\}$ from
$\{G_0, G_1, \ldots\}$ along with the length $L$ sparse coefficient vectors
located at the center of each sampled patch, $\{z_0, z_1, \ldots\}$.  We
rectify each of these sparse vectors and append a constant term:
\begin{equation}
   \hat{z}_i = \left[~~\max(z_i^T,0),~~\max(-z_i^T,0),~~1~~\right]^T
   \label{eq:rectification}
\end{equation}
Our patch-level transfer learning problem is now to find $F(\cdot)$ such that:
\begin{equation}
   F(\hat{z}_i) \approx g_i \quad \forall i
\end{equation}
where $\hat{z}_i \in \Re^{2L+1}$ is a vector of sparse coefficients and
$g_i \in \Re^{m \times m \times h}$ is a target ground-truth patch.  Here,
$h$ denotes the number of channels in the ground-truth (and its predicted
reconstruction).

While one could still choose any method for modeling $F(\cdot)$, we make an
extremely simple and efficient choice, with the expectation that the sparse
representation will be rich enough that simple transfer functions will work
well.  Specifically, we split $F(\cdot)$ into a set
$[f_0, f_1, ..., f_{(m^2h-1)}]$ of independently trained predictors $f(\cdot)$,
one for each of the $m^2h$ elements of the output patch.  Our transfer
learning problem is now:
\begin{equation}
   f_j(\hat{z}_i) \approx g_i[j] \quad \forall i,j
\end{equation}
As all experiments in this paper deal with ground-truth in the form of binary
indicator vectors, we set each $f_j(\cdot)$ to be a logistic classifier and
train its coefficients using L2-regularized logistic regression.

Predicting $m \times m$ patches means that each element of the output
reconstruction is an average of outputs from $m^2$ different $f_j(\cdot)$
classifiers.  Moreover, one would expect (and we observe in practice) the
accuracy of the classifiers to be spatially varying.  Predicted labels of
pixels more distant from the patch center are less reliable than those
nearby.  To correct for this, we weight predicted patches with a Gaussian
kernel when spatially averaging them during reconstruction.

\begin{figure*}[t!]
   \begin{center}
      \setlength\fboxsep{0pt}
      \begin{minipage}[t]{0.19\linewidth}
         \fbox{\includegraphics[width=1.00\linewidth]{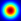}}
      \end{minipage}
      \begin{minipage}[t]{0.19\linewidth}
         \fbox{\includegraphics[width=1.00\linewidth]{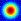}}
      \end{minipage}
      \begin{minipage}[t]{0.19\linewidth}
         \fbox{\includegraphics[width=1.00\linewidth]{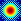}}
      \end{minipage}
      \begin{minipage}[t]{0.19\linewidth}
         \fbox{\includegraphics[width=1.00\linewidth]{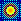}}
      \end{minipage}
      \begin{minipage}[t]{0.19\linewidth}
         \fbox{\includegraphics[width=1.00\linewidth]{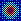}}
      \end{minipage} \\
      \begin{minipage}[t]{0.19\linewidth}
         \begin{center}
            \scriptsize{\textbf{\textsf{\#nonzeros = 441}}}
         \end{center}
      \end{minipage}
      \begin{minipage}[t]{0.19\linewidth}
         \begin{center}\scriptsize{\textbf{\textsf{389}}}\end{center}
      \end{minipage}
      \begin{minipage}[t]{0.19\linewidth}
         \begin{center}\scriptsize{\textbf{\textsf{289}}}\end{center}
      \end{minipage}
      \begin{minipage}[t]{0.19\linewidth}
         \begin{center}\scriptsize{\textbf{\textsf{205}}}\end{center}
      \end{minipage}
      \begin{minipage}[t]{0.19\linewidth}
         \begin{center}\scriptsize{\textbf{\textsf{157}}}\end{center}
      \end{minipage}
   \end{center}
   \vspace{-0.04\linewidth}
   \caption{
      \textbf{Subsampled patch averaging kernels.}
         Instead of uniformly averaging overlapping patch predictions during
         reconstruction, we weight them with a Gaussian kernel to model their
         spatially varying reliability.  As a classifier evaluation can be
         skipped if its prediction will receive zero weight within the
         averaging procedure, we adaptively sparsify the kernel to achieve a
         runtime speedup.  Using the aggressively subsampled \emph{(rightmost)}
         kernel is $3$x faster than using the non-subsampled \emph{(leftmost)}
         version, and offers equivalent accuracy.  We also save the expense of
         training unused classifiers.
   }
   \label{fig:focus}
\end{figure*}

Additionally, we would like the computation time for prediction to grow
more slowly than $O(m^2)$ as patch size increases.  Because predictions
originating from similar spatial locations are likely to be correlated and a
Gaussian kernel gives distant neighbors small weight, we construct an adaptive
kernel $\mathcal{W}$, which approximates the Gaussian, taking fewer samples
with increasing distance, but upweighting them to compensate for decreased
sample density.  Specifically:
\begin{equation}
   \mathcal{W}(x,y;\sigma) =
      \left\{
         \begin{array}{ll}
            \mathcal{G}(x,y;\sigma) / \rho(x,y)
               & \mbox{\quad if $(x,y) \in \mathcal{S}$}\\
            0  & \mbox{\quad otherwise}
         \end{array}
      \right.
\end{equation}
where $\mathcal{G}$ is a 2D Gaussian, $\mathcal{S}$ is a set of sample points,
and $\rho(x,y)$ measures the local density of sample points.  Figure
\ref{fig:focus} provides an illustration of $\mathcal{W}$ for fixed~$\sigma$
and sampling patterns which repeatedly halve density at various radii.

We report all experimental results using the adaptively sampled approximate
Gaussian kernel during reconstruction.  We found it to perform equivalently to
the full Gaussian kernel and better than uniform patch weighting.  The
adaptive weight kernel not only reduces runtime, but also reduces training
time as we neither run nor train the $f_j(\cdot)$ classifiers that the kernel
assigns zero weight.

\section{Multipath Network}
\label{sec:multipath}

Sections~\ref{sec:representation} and~\ref{sec:transfer} describe our
system for reconstructive sparse code transfer in the context of a single
generatively learned patch dictionary $D$ and the resulting sparse
representation.  In practice, we must offer the system a richer view of the
input than can be obtained from coding against a single dictionary.  To
accomplish this, we borrow the multipath sparse coding framework of
Bo~\emph{et al.}~\cite{MHMP} which combines two strategies for building richer
representations.

First, the image is rescaled and all scales are coded against multiple
dictionaries for patches of varying size.  Second, the output sparse
representation is pooled, subsampled, and then treated as a new input signal
for another layer of sparse coding.  Figure~\ref{fig:network} describes the
network architecture we have chosen in order to implement this strategy.  We
use rectification followed by hybrid average-max pooling (the average of
nonzero coefficients) between layers 1 and 2.  For $5\times5$ patches, we pool
over $3\times3$ windows and subsample by a factor of $2$, while for
$11\times11$ patches, we pool over $5\times5$ windows and subsample by a
factor of $4$.

We concatenate all representations generated by the $512$-atom dictionaries,
rectify, and upsample them so that they live on the original image grid.
This results in a $36000$-dimensional sparse vector representation for each
image pixel.  Despite the high dimensionality, there are only a few hundred
nonzero entries per pixel, so total computational work is quite reasonable.

The dictionary transfer stage described in Section~\ref{sec:transfer} now
operates on these high-dimensional concatenated sparse vectors ($L=36000$)
instead of the output of a single dictionary.  Training is more expensive,
but classification and reconstruction is still cheap.  The cost of evaluating
the logistic classifiers scales with the number of nonzero coefficients in
the sparse representations rather than the dimensionality.  As a speedup for
training, we drop a different random $50\%$ of the representation for
each of the logistic classifiers.

\section{Experiments}
\label{sec:experiments}

We apply multipath reconstructive sparse code transfer to two pixel labeling
tasks: contour detection on the Berkeley segmentation dataset
(BSDS)~\cite{BSDS}, and semantic labeling of human faces (into skin, hair, and
background) on the part subset~\cite{LFW-parts} of the Labeled Faces in the
Wild (LFW) dataset~\cite{LFW}.  We use the network structure in
Figure~\ref{fig:network} in both sets of experiments, with the only
difference being that we apply a zero-mean transform to patch channels
prior to encoding in the BSDS experiments.  This choice was simply made to
increase dictionary efficiency in the case of contour detection, where
absolute color is likely less important.  For experiments on the LFW dataset,
we directly encode raw patches.

\subsection{Contour Detection}
\label{sec:contours}

\begin{figure*}
   \begin{center}
      \setlength\fboxsep{0pt}
      \fbox{\includegraphics[width=0.115\linewidth]{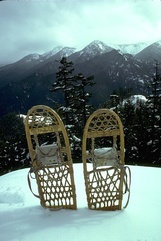}}
      \fbox{\includegraphics[width=0.115\linewidth]{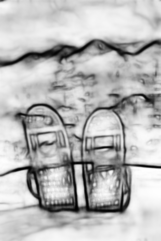}}
      \hfill
      \fbox{\includegraphics[width=0.115\linewidth]{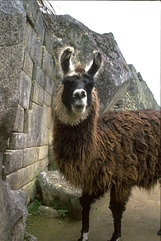}}
      \fbox{\includegraphics[width=0.115\linewidth]{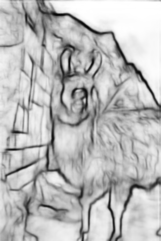}}
      \hfill
      \fbox{\includegraphics[width=0.115\linewidth]{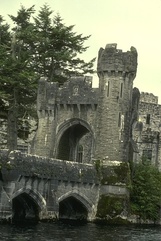}}
      \fbox{\includegraphics[width=0.115\linewidth]{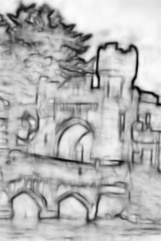}}
      \hfill
      \fbox{\includegraphics[width=0.115\linewidth]{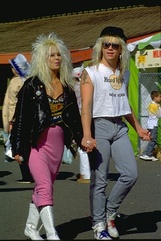}}
      \fbox{\includegraphics[width=0.115\linewidth]{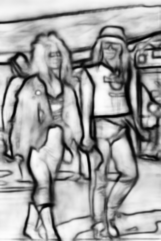}}\\
      \vspace{0.01\linewidth}
      \fbox{\includegraphics[width=0.115\linewidth]{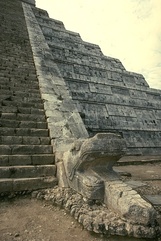}}
      \fbox{\includegraphics[width=0.115\linewidth]{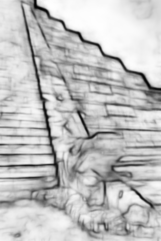}}
      \hfill
      \fbox{\includegraphics[width=0.115\linewidth]{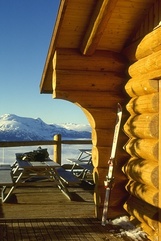}}
      \fbox{\includegraphics[width=0.115\linewidth]{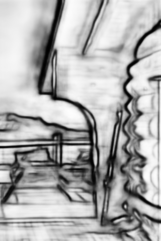}}
      \hfill
      \fbox{\includegraphics[width=0.115\linewidth]{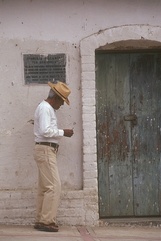}}
      \fbox{\includegraphics[width=0.115\linewidth]{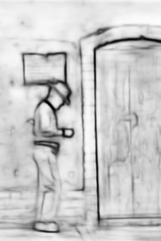}}
      \hfill
      \fbox{\includegraphics[width=0.115\linewidth]{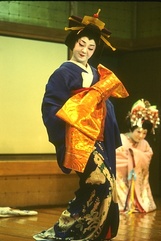}}
      \fbox{\includegraphics[width=0.115\linewidth]{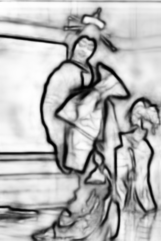}}\\
      \vspace{0.01\linewidth}
      \fbox{\includegraphics[width=0.115\linewidth]{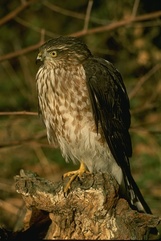}}
      \fbox{\includegraphics[width=0.115\linewidth]{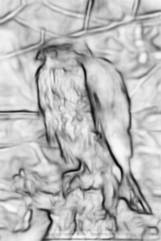}}
      \hfill
      \fbox{\includegraphics[width=0.115\linewidth]{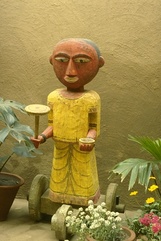}}
      \fbox{\includegraphics[width=0.115\linewidth]{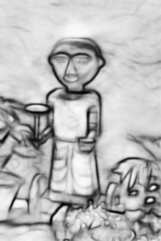}}
      \hfill
      \fbox{\includegraphics[width=0.115\linewidth]{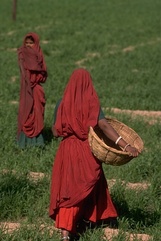}}
      \fbox{\includegraphics[width=0.115\linewidth]{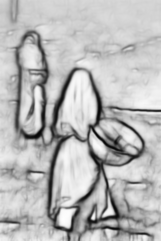}}
      \hfill
      \fbox{\includegraphics[width=0.115\linewidth]{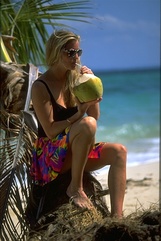}}
      \fbox{\includegraphics[width=0.115\linewidth]{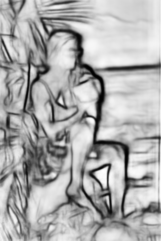}}\\
      \vspace{0.01\linewidth}
      \fbox{\includegraphics[width=0.115\linewidth]{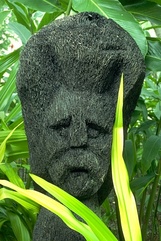}}
      \fbox{\includegraphics[width=0.115\linewidth]{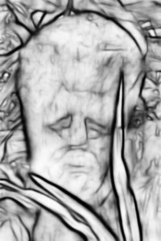}}
      \hfill
      \fbox{\includegraphics[width=0.115\linewidth]{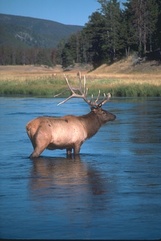}}
      \fbox{\includegraphics[width=0.115\linewidth]{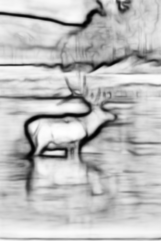}}
      \hfill
      \fbox{\includegraphics[width=0.115\linewidth]{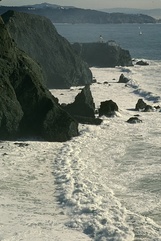}}
      \fbox{\includegraphics[width=0.115\linewidth]{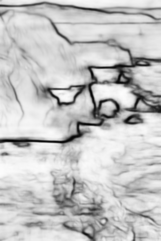}}
      \hfill
      \fbox{\includegraphics[width=0.115\linewidth]{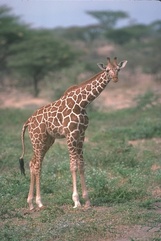}}
      \fbox{\includegraphics[width=0.115\linewidth]{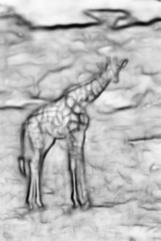}}\\
      \vspace{0.01\linewidth}
      \fbox{\includegraphics[width=0.115\linewidth]{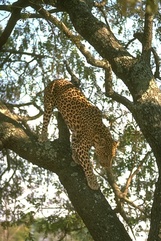}}
      \fbox{\includegraphics[width=0.115\linewidth]{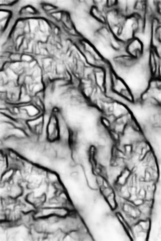}}
      \hfill
      \fbox{\includegraphics[width=0.115\linewidth]{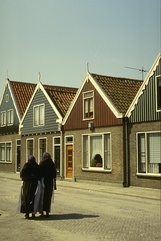}}
      \fbox{\includegraphics[width=0.115\linewidth]{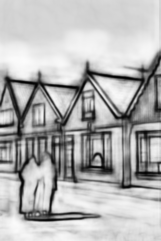}}
      \hfill
      \fbox{\includegraphics[width=0.115\linewidth]{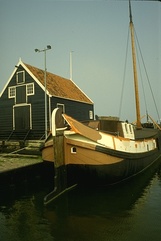}}
      \fbox{\includegraphics[width=0.115\linewidth]{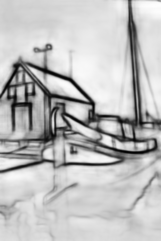}}
      \hfill
      \fbox{\includegraphics[width=0.115\linewidth]{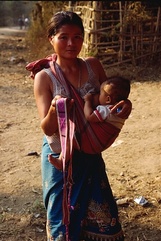}}
      \fbox{\includegraphics[width=0.115\linewidth]{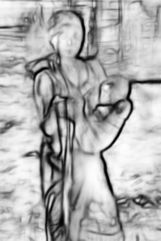}}\\
      \vspace{0.01\linewidth}
      \fbox{\includegraphics[width=0.115\linewidth]{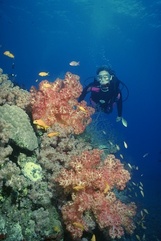}}
      \fbox{\includegraphics[width=0.115\linewidth]{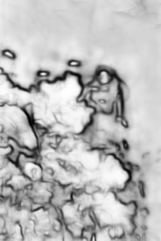}}
      \hfill
      \fbox{\includegraphics[width=0.115\linewidth]{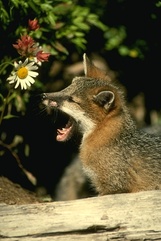}}
      \fbox{\includegraphics[width=0.115\linewidth]{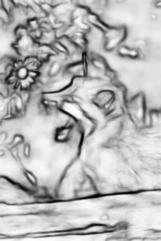}}
      \hfill
      \fbox{\includegraphics[width=0.115\linewidth]{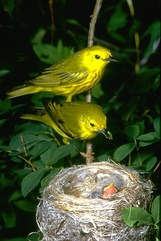}}
      \fbox{\includegraphics[width=0.115\linewidth]{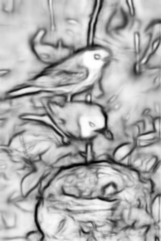}}
      \hfill
      \fbox{\includegraphics[width=0.115\linewidth]{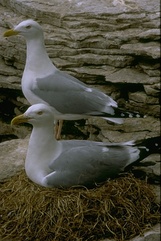}}
      \fbox{\includegraphics[width=0.115\linewidth]{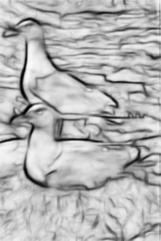}}\\
      \vspace{0.01\linewidth}
      \fbox{\includegraphics[width=0.115\linewidth]{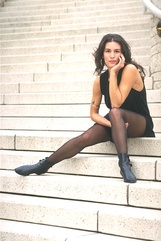}}
      \fbox{\includegraphics[width=0.115\linewidth]{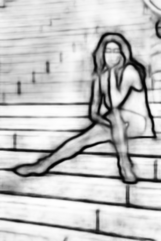}}
      \hfill
      \fbox{\includegraphics[width=0.115\linewidth]{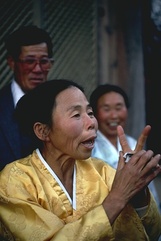}}
      \fbox{\includegraphics[width=0.115\linewidth]{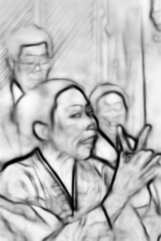}}
      \hfill
      \fbox{\includegraphics[width=0.115\linewidth]{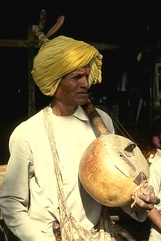}}
      \fbox{\includegraphics[width=0.115\linewidth]{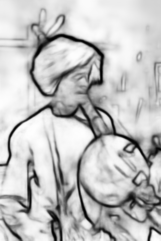}}
      \hfill
      \fbox{\includegraphics[width=0.115\linewidth]{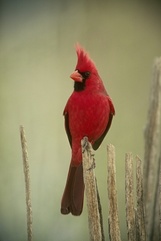}}
      \fbox{\includegraphics[width=0.115\linewidth]{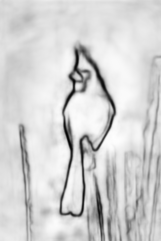}}\\
      \vspace{0.01\linewidth}
      \fbox{\includegraphics[width=0.115\linewidth]{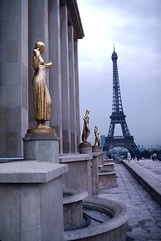}}
      \fbox{\includegraphics[width=0.115\linewidth]{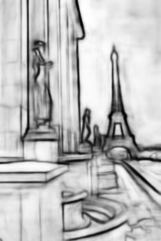}}
      \hfill
      \fbox{\includegraphics[width=0.115\linewidth]{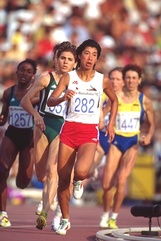}}
      \fbox{\includegraphics[width=0.115\linewidth]{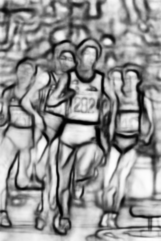}}
      \hfill
      \fbox{\includegraphics[width=0.115\linewidth]{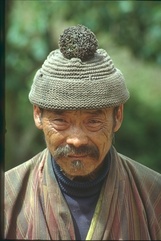}}
      \fbox{\includegraphics[width=0.115\linewidth]{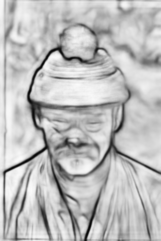}}
      \hfill
      \fbox{\includegraphics[width=0.115\linewidth]{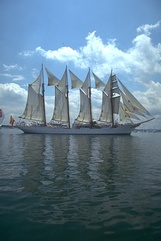}}
      \fbox{\includegraphics[width=0.115\linewidth]{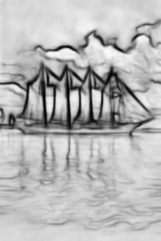}}
   \end{center}
   \vspace{-0.05\linewidth}
   \caption{
      \textbf{Contour detection results on BSDS500.}
         We show images and corresponding contours produced via reconstructive
         sparse code transfer.  Contours are displayed prior to applying
         non-maximal suppression and thinning for benchmarking purposes.
   }
   \label{fig:contour_results}
\end{figure*}

\begin{figure*}
   \begin{center}
      \includegraphics[width=0.49\linewidth, clip=true, trim=1.5in 2.75in 1.25in 2.725in]{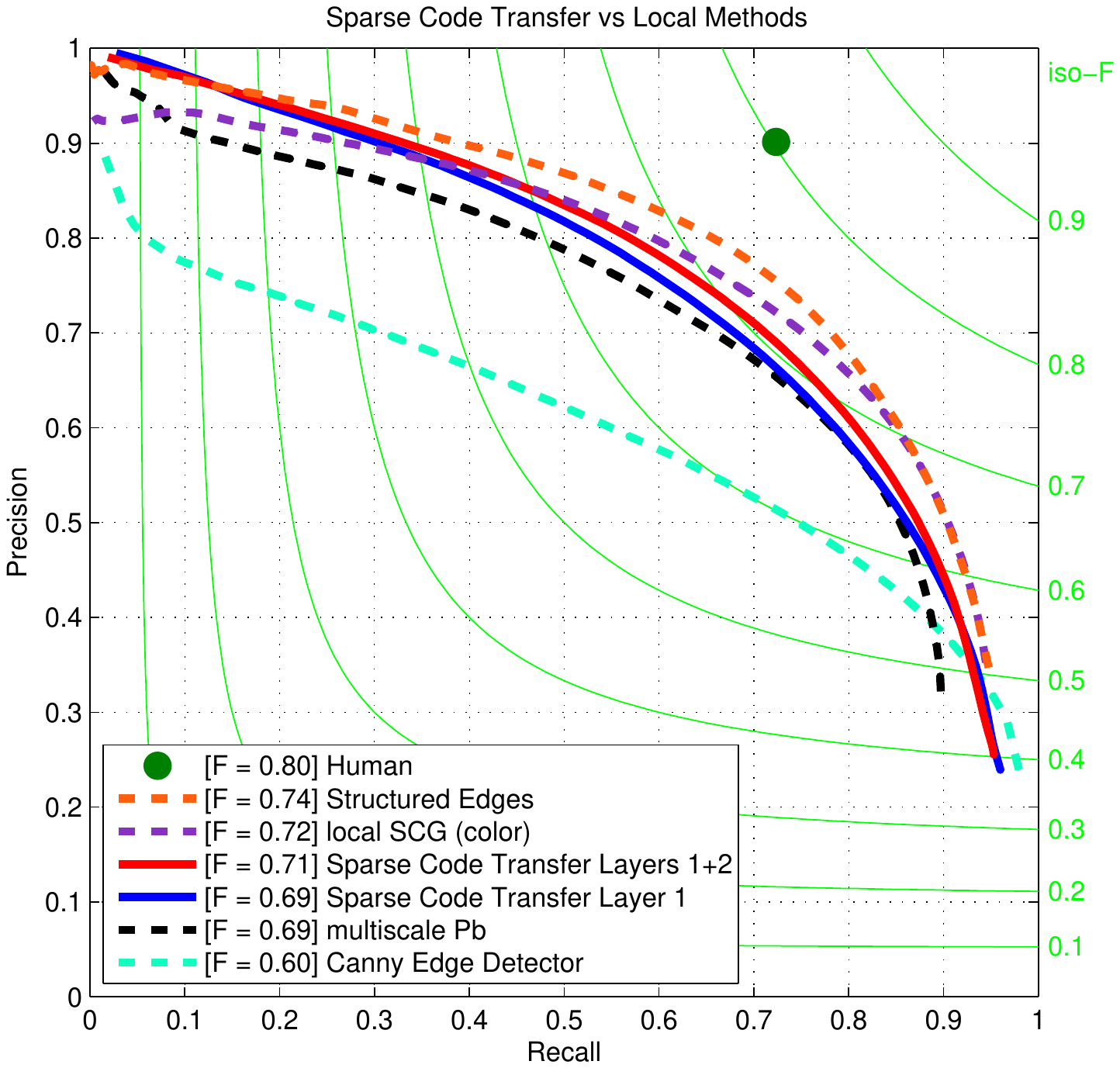}
      \hfill
      \includegraphics[width=0.49\linewidth, clip=true, trim=1.5in 2.75in 1.25in 2.725in]{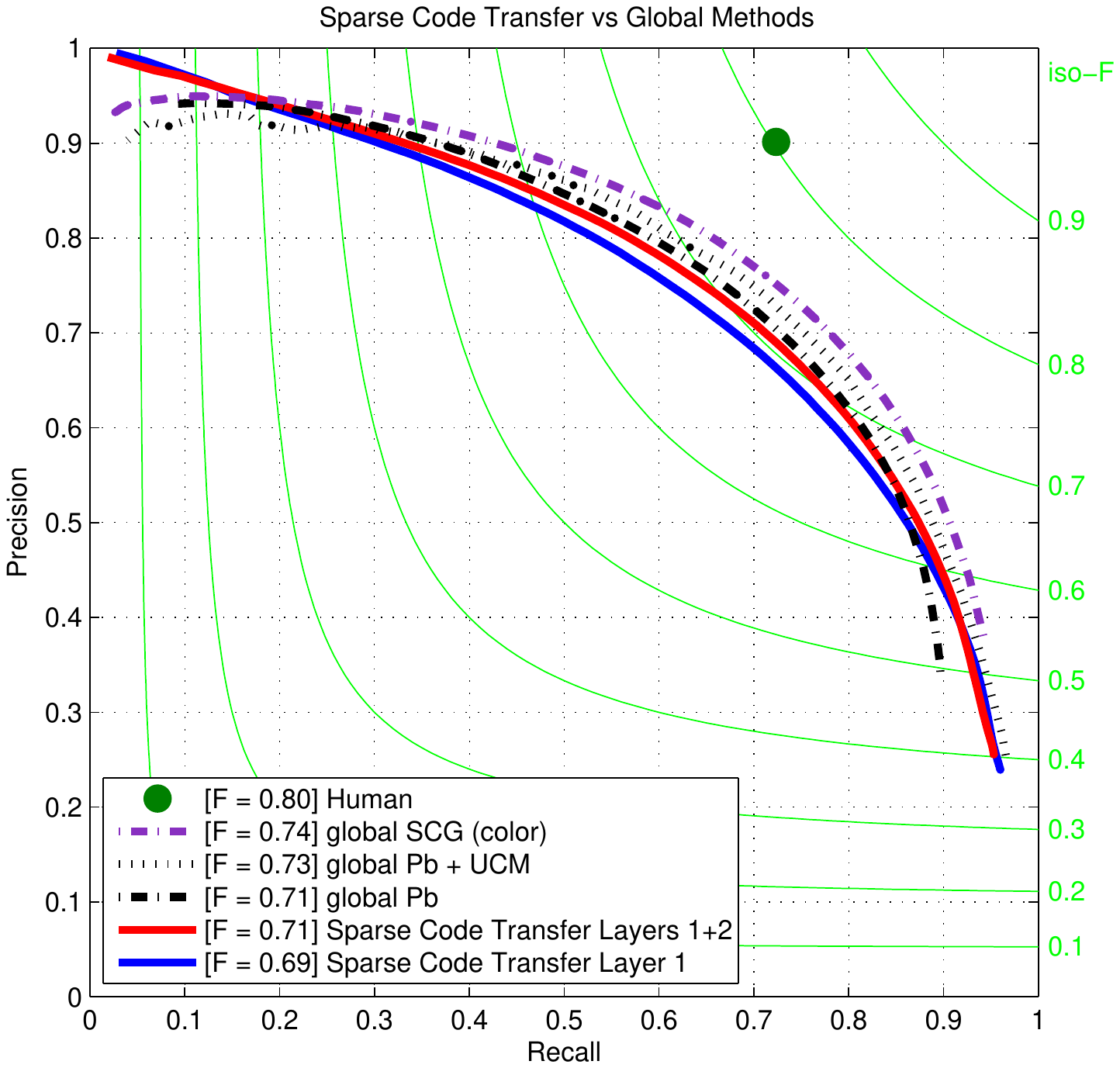}
   \end{center}
   \vspace{-0.05\linewidth}
   \caption{
      \textbf{Contour detection performance on BSDS500.}
         Our contour detector (solid red curve) achieves a maximum F-measure
         $(\frac{2 \cdot Precision \cdot Recall}{Precision + Recall})$ of
         $0.71$, similar to other leading approaches.  Table
         \ref{tab:contour_benchmark} elaborates with more performance metrics.
      \emph{Left:}
         We show full precision-recall curves for algorithms that, like ours,
         predict boundary strength directly from local image patches.  Of
         these algorithms, sparse code transfer is the only one free of
         reliance on hand-designed features or filters.  Note that addition
         of the second layer improves our system's performance, as seen in
         the jump from the blue to red curve.
      \emph{Right:}
         Post-processing steps that perform global reasoning on top of locally
         detected contours can further boost performance.  Application of
         spectral clustering to multiscale Pb and local SCG yields superior
         results shown as global Pb and global SCG, respectively.  Further
         transforming global Pb via an Ultrametric Contour Map (UCM)~\cite{UCM}
         yields an additional boost.  Without any such post-processing, our
         local detector offers performance equivalent to that of global Pb.
   }
   \label{fig:contour_benchmark}
\end{figure*}

Figure~\ref{fig:contour_results} shows contour detection results on example
images from the test set of the 500 image version~\cite{gPb-UCM} of the
BSDS~\cite{BSDS}.  Figure~\ref{fig:contour_benchmark} shows the
precision-recall curve for our contour detector as benchmarked against
human-drawn ground-truth.  Performance is comparable to the heavily-engineered
state-of-the-art global Pb (gPb) detector~\cite{gPb-UCM}.

Note that both gPb and SCG~\cite{SCG} apply a spectral clustering procedure on
top of their detector output in order to generate a cleaner globally
consistent result.  In both cases, this extra step provides a performance
boost.  Table~\ref{tab:contour_benchmark} displays a more nuanced comparison
of our contour detection performance with that of SCG before globalization.
Our detector performs comparably to (local) SCG.  We expect that inclusion of
a sophisticated spectral integration step~\cite{MYP:ICCV:2013} will further
boost our contour detection performance, but leave the proof to future work.

It is also worth emphasizing that our system is the only method in
Table~\ref{tab:contour_benchmark} that relies on neither hand-crafted
filters (global Pb, SCG) nor hand-crafted features (global Pb,
Structured Edges).  Our system is learned entirely from data and even relies
on a \emph{generatively trained} representation as a critical component.

\begin{figure*}
   \begin{center}
      \hfill
      \begin{minipage}[t]{0.572\linewidth}
         \vspace{0pt}
         \setlength\fboxsep{0pt}
         \fbox{\includegraphics[width=0.31\linewidth]{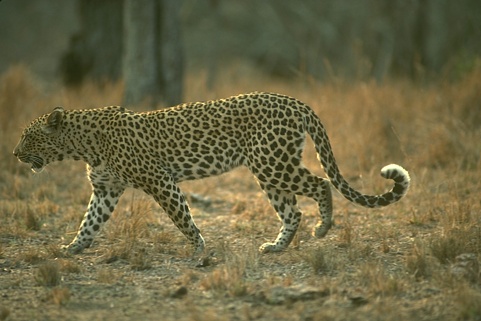}}
         \fbox{\includegraphics[width=0.31\linewidth]{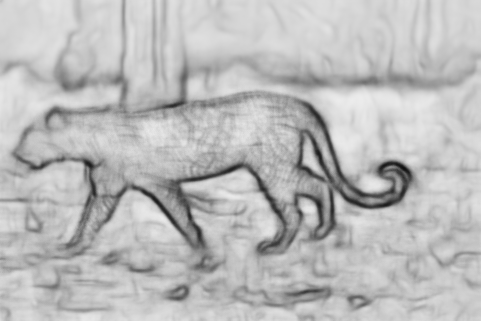}}
         \fbox{\includegraphics[width=0.31\linewidth]{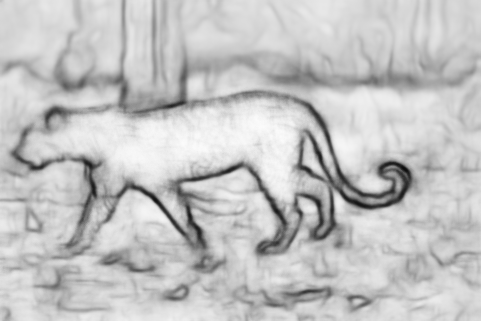}}
         \vspace{0.01\linewidth}\\
         \fbox{\includegraphics[width=0.31\linewidth]{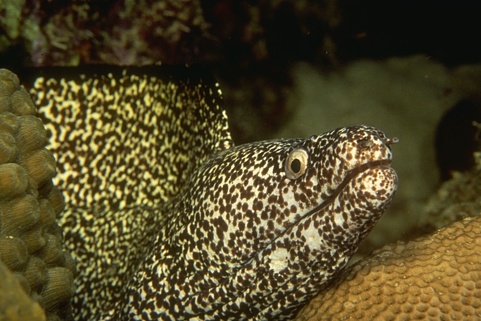}}
         \fbox{\includegraphics[width=0.31\linewidth]{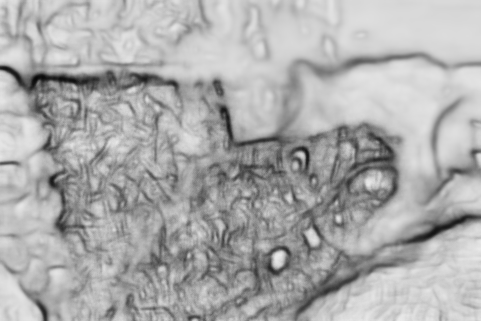}}
         \fbox{\includegraphics[width=0.31\linewidth]{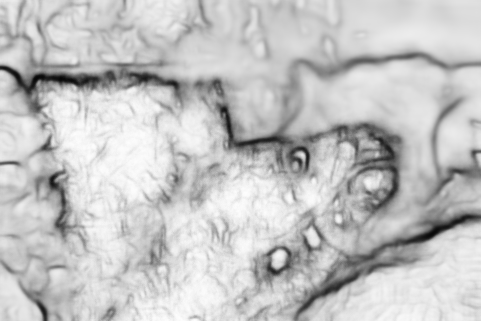}}
         \vspace{0.01\linewidth}\\
         \fbox{\includegraphics[width=0.31\linewidth]{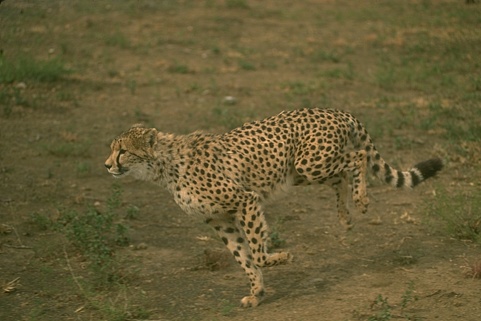}}
         \fbox{\includegraphics[width=0.31\linewidth]{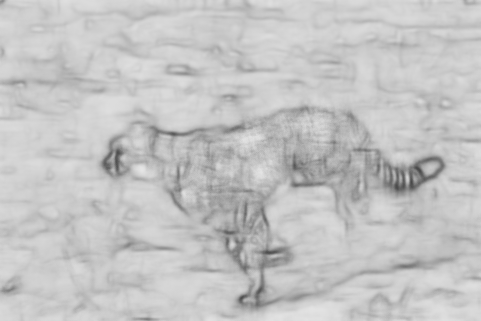}}
         \fbox{\includegraphics[width=0.31\linewidth]{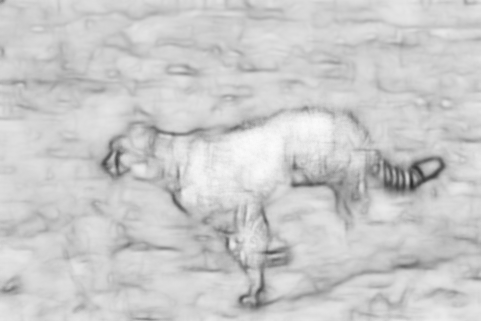}}
      \end{minipage}
      \begin{minipage}[t]{0.41\linewidth}
         \vspace{0pt}
         \setlength\fboxsep{0pt}
         \fbox{\includegraphics[width=0.30\linewidth]{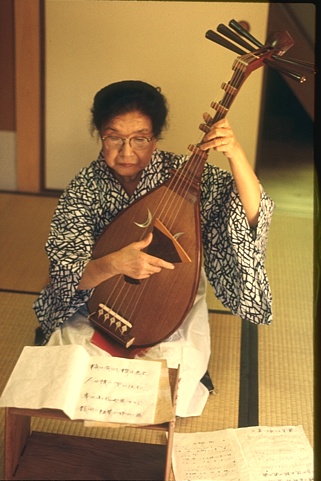}}
         \fbox{\includegraphics[width=0.30\linewidth]{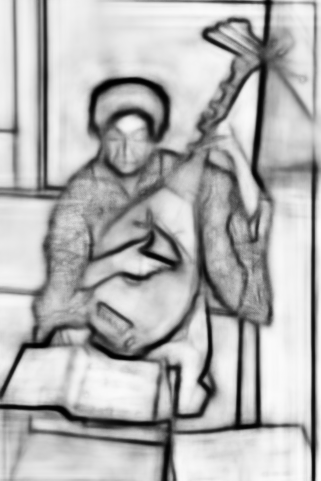}}
         \fbox{\includegraphics[width=0.30\linewidth]{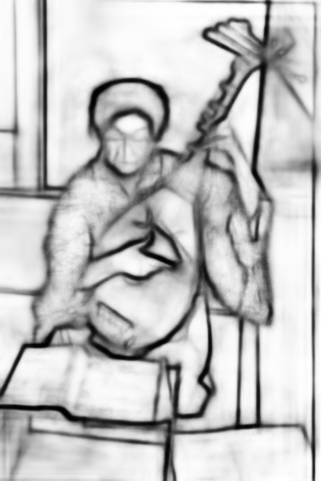}}
         \vspace{0.01\linewidth}\\
         \fbox{\includegraphics[width=0.30\linewidth]{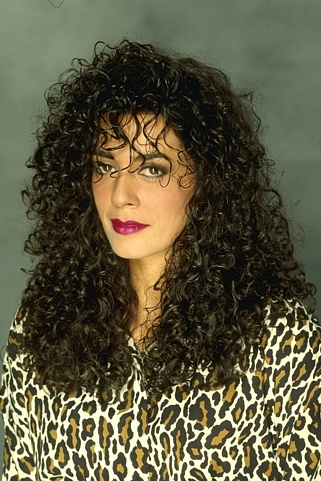}}
         \fbox{\includegraphics[width=0.30\linewidth]{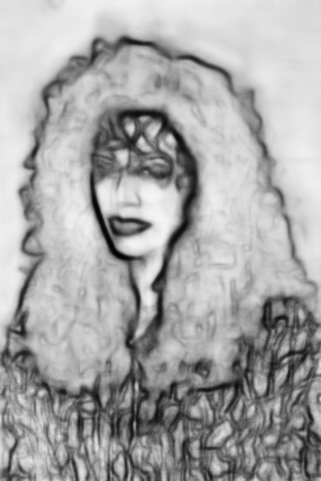}}
         \fbox{\includegraphics[width=0.30\linewidth]{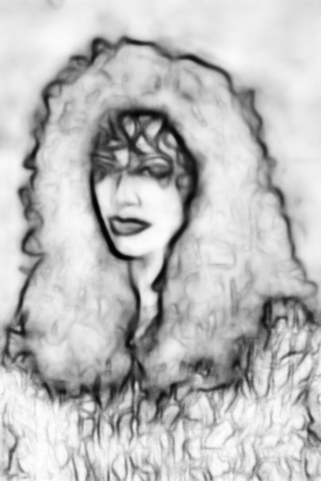}}
      \end{minipage}
      \hfill \\
      \vspace{0.0075\linewidth}
      \hfill
      \begin{minipage}[t]{0.572\linewidth}
         \begin{center}
         \begin{minipage}[t]{0.30\linewidth}\hfill\end{minipage}
         \begin{minipage}[t]{0.30\linewidth}
            \begin{center}
               \scriptsize{\textbf{\textsf{Layer 1}}}
               \hspace{0.15\linewidth}
            \end{center}
         \end{minipage}
         \begin{minipage}[t]{0.30\linewidth}
            \begin{center}\scriptsize{\textbf{\textsf{Layers 1+2}}}\end{center}
         \end{minipage}
         \end{center}
      \end{minipage}
      \begin{minipage}[t]{0.41\linewidth}
         \begin{center}
         \begin{minipage}[t]{0.30\linewidth}\hfill\end{minipage}
         \begin{minipage}[t]{0.28\linewidth}
            \begin{center}\scriptsize{\textbf{\textsf{~Layer 1}}}\end{center}
         \end{minipage}
         \begin{minipage}[t]{0.35\linewidth}
            \begin{center}\scriptsize{\textbf{\textsf{Layers 1+2}}}\end{center}
         \end{minipage}
         \end{center}
      \end{minipage}
      \hfill
   \end{center}
   \vspace{-0.04\linewidth}
   \caption{
      \textbf{Texture understanding and network depth.}
         From left to right, we display an image, the contours detected using
         only the sparse representation from the layer~$1$ dictionaries in
         Figure~\ref{fig:network}, and the contours detected using the
         representation from both layers~$1$ and~$2$.  Inclusion of the second
         layer is crucial to enabling the system to suppress undesirable
         fine-scale texture edges.
   }
   \label{fig:texture}
\end{figure*}

Additional analysis of our results yields the interesting observation
that the second layer of our multipath network appears crucial to texture
understanding.  Figure~\ref{fig:texture} shows a comparison of contour
detection results when our system is restricted to use only layer 1 versus
results when the system uses the sparse representation from both layers
1 and 2.  Inclusion of the second layer (deep sparse coding) essentially
allows the classification stage to learn an off switch for texture edges.

\begin{table}
   \begin{center}
   \begin{scriptsize}
   \begin{tabular}{|l|ccc|cc|c|}
      \cline{2-7}
      \multicolumn{1}{l|}{}              & \multicolumn{3}{c|}{Performance Metric}      & \multicolumn{2}{c|}{Hand-Designed} & Spectral \\
      \multicolumn{1}{l|}{}              & ODS F         & OIS F         & AP           & Features? & Filters?               & Globalization? \\
      \hline
      Human                              & $0.80$        & $0.80$        & $-$          & $-$       & $-$                    & $-$ \\
      \hline
      {Structured Edges~\cite{SE}}       & $\bf{0.74}$   & $\bf{0.76}$   & $\bf{0.78}$  & yes       & no                     & no \\
      {local SCG (color)~\cite{SCG}}     & $0.72$        & $0.74$        & $0.75$       & no        & yes                    & no \\
      {Sparse Code Transfer Layers 1+2~} & $0.71$        & $0.72$        & $0.74$       & \bf{no}   & \bf{no}                & no \\
      {Sparse Code Transfer Layer 1}     & $0.69$        & $0.71$        & $0.72$       & \bf{no}   & \bf{no}                & no \\
      {local SCG (gray)~\cite{SCG}}      & $0.69$        & $0.71$        & $0.71$       & no        & yes                    & no \\
      {multiscale Pb~\cite{gPb-UCM}}     & $0.69$        & $0.71$        & $0.68$       & yes       & yes                    & no \\
      {Canny Edge Detector~\cite{Canny}} & $0.60$        & $0.63$        & $0.58$       & yes       & yes                    & no \\
      \hline
      {global SCG (color)~\cite{SCG}}    & $\bf{0.74}$   & $\bf{0.76}$   & $0.77$       & yes       & yes                    & yes \\
      {global Pb + UCM~\cite{gPb-UCM}}   & $0.73$        & $\bf{0.76}$   & $0.73$       & yes       & yes                    & yes + UCM  \\
      {global Pb~\cite{gPb-UCM}}         & $0.71$        & $0.74$        & $0.65$       & yes       & yes                    & yes \\
      \hline
   \end{tabular}
   \end{scriptsize}
   \end{center}
   \caption{
      \textbf{Contour benchmarks on BSDS500.}
      Performance of our sparse code transfer technique is competitive with
      the current best performing contour detection systems
      \cite{gPb-UCM,SCG,SE}.  Shown are the detector F-measures when choosing
      an optimal threshold for the entire dataset (ODS) or per image (OIS), as
      well as the average precision (AP).  The upper block of the table reports
      scores prior to application of spectral globalization, while the lower
      block reports improved results of some systems afterwards.  Note that
      our system is the only approach in which both the feature representation
      and classifier are entirely learned.
   }
   \label{tab:contour_benchmark}
\end{table}

\subsection{Semantic Labeling of Faces}
\label{sec:faces}

\begin{figure*}
   \begin{center}
      \setlength\fboxsep{0pt}
      \fbox{\includegraphics[width=0.10\linewidth]{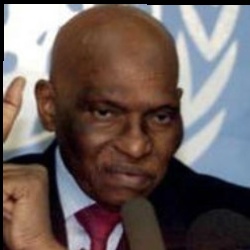}}
      \fbox{\includegraphics[width=0.10\linewidth]{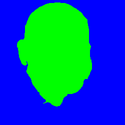}}
      \fbox{\includegraphics[width=0.10\linewidth]{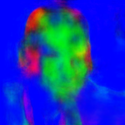}}
      \hfill
      \fbox{\includegraphics[width=0.10\linewidth]{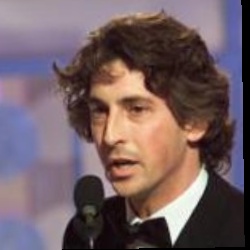}}
      \fbox{\includegraphics[width=0.10\linewidth]{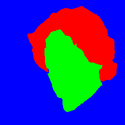}}
      \fbox{\includegraphics[width=0.10\linewidth]{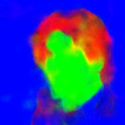}}
      \hfill
      \fbox{\includegraphics[width=0.10\linewidth]{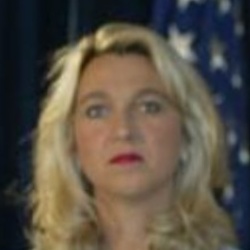}}
      \fbox{\includegraphics[width=0.10\linewidth]{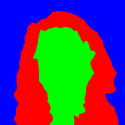}}
      \fbox{\includegraphics[width=0.10\linewidth]{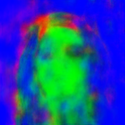}}\\
      \vspace{0.01\linewidth}
      \fbox{\includegraphics[width=0.10\linewidth]{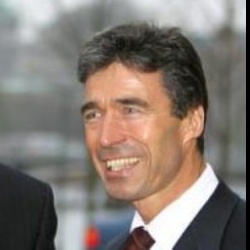}}
      \fbox{\includegraphics[width=0.10\linewidth]{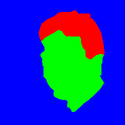}}
      \fbox{\includegraphics[width=0.10\linewidth]{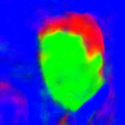}}
      \hfill
      \fbox{\includegraphics[width=0.10\linewidth]{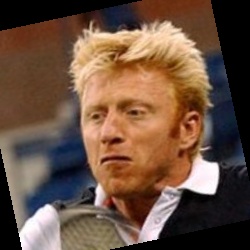}}
      \fbox{\includegraphics[width=0.10\linewidth]{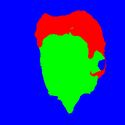}}
      \fbox{\includegraphics[width=0.10\linewidth]{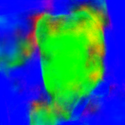}}
      \hfill
      \fbox{\includegraphics[width=0.10\linewidth]{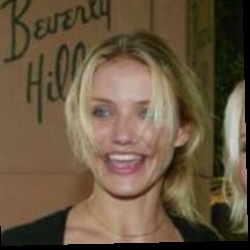}}
      \fbox{\includegraphics[width=0.10\linewidth]{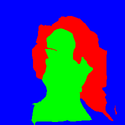}}
      \fbox{\includegraphics[width=0.10\linewidth]{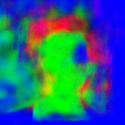}}\\
      \hfill
      \fbox{\includegraphics[width=0.10\linewidth]{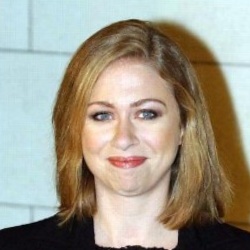}}
      \fbox{\includegraphics[width=0.10\linewidth]{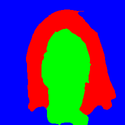}}
      \fbox{\includegraphics[width=0.10\linewidth]{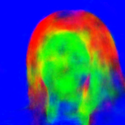}}
      \hfill
      \fbox{\includegraphics[width=0.10\linewidth]{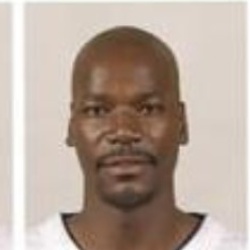}}
      \fbox{\includegraphics[width=0.10\linewidth]{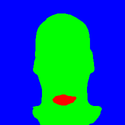}}
      \fbox{\includegraphics[width=0.10\linewidth]{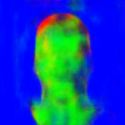}}
      \hfill
      \fbox{\includegraphics[width=0.10\linewidth]{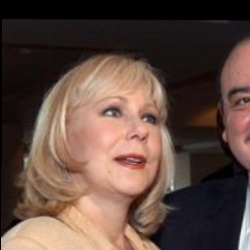}}
      \fbox{\includegraphics[width=0.10\linewidth]{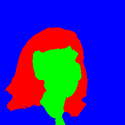}}
      \fbox{\includegraphics[width=0.10\linewidth]{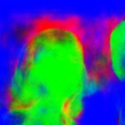}}\\
      \vspace{0.01\linewidth}
      \fbox{\includegraphics[width=0.10\linewidth]{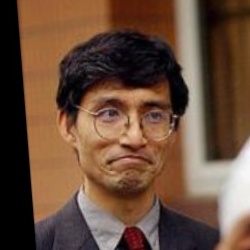}}
      \fbox{\includegraphics[width=0.10\linewidth]{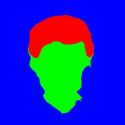}}
      \fbox{\includegraphics[width=0.10\linewidth]{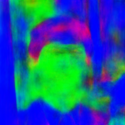}}
      \hfill
      \fbox{\includegraphics[width=0.10\linewidth]{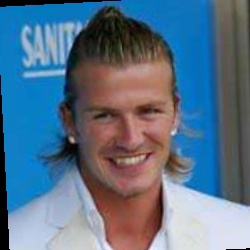}}
      \fbox{\includegraphics[width=0.10\linewidth]{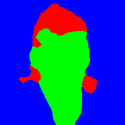}}
      \fbox{\includegraphics[width=0.10\linewidth]{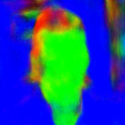}}
      \hfill
      \fbox{\includegraphics[width=0.10\linewidth]{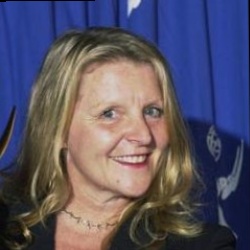}}
      \fbox{\includegraphics[width=0.10\linewidth]{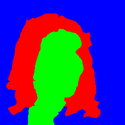}}
      \fbox{\includegraphics[width=0.10\linewidth]{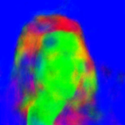}}\\
      \vspace{0.01\linewidth}
      \fbox{\includegraphics[width=0.10\linewidth]{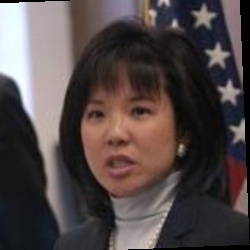}}
      \fbox{\includegraphics[width=0.10\linewidth]{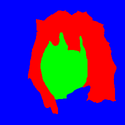}}
      \fbox{\includegraphics[width=0.10\linewidth]{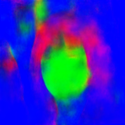}}
      \hfill
      \fbox{\includegraphics[width=0.10\linewidth]{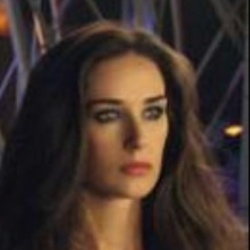}}
      \fbox{\includegraphics[width=0.10\linewidth]{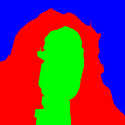}}
      \fbox{\includegraphics[width=0.10\linewidth]{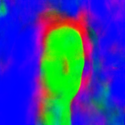}}
      \hfill
      \fbox{\includegraphics[width=0.10\linewidth]{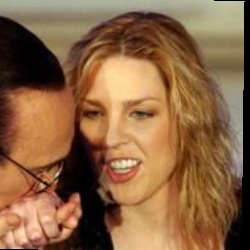}}
      \fbox{\includegraphics[width=0.10\linewidth]{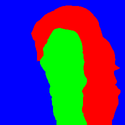}}
      \fbox{\includegraphics[width=0.10\linewidth]{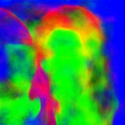}}\\
      \vspace{0.01\linewidth}
      \fbox{\includegraphics[width=0.10\linewidth]{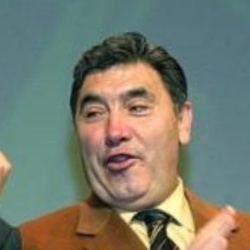}}
      \fbox{\includegraphics[width=0.10\linewidth]{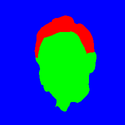}}
      \fbox{\includegraphics[width=0.10\linewidth]{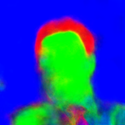}}
      \hfill
      \fbox{\includegraphics[width=0.10\linewidth]{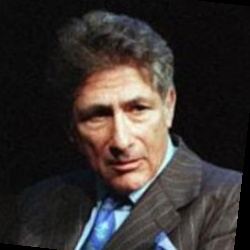}}
      \fbox{\includegraphics[width=0.10\linewidth]{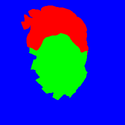}}
      \fbox{\includegraphics[width=0.10\linewidth]{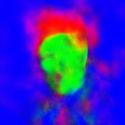}}
      \hfill
      \fbox{\includegraphics[width=0.10\linewidth]{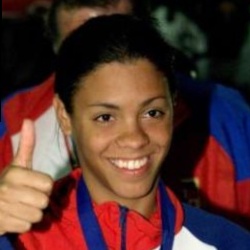}}
      \fbox{\includegraphics[width=0.10\linewidth]{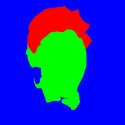}}
      \fbox{\includegraphics[width=0.10\linewidth]{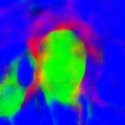}}\\
      \vspace{0.01\linewidth}
      \fbox{\includegraphics[width=0.10\linewidth]{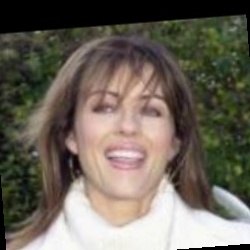}}
      \fbox{\includegraphics[width=0.10\linewidth]{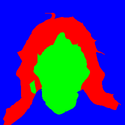}}
      \fbox{\includegraphics[width=0.10\linewidth]{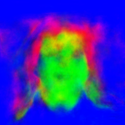}}
      \hfill
      \fbox{\includegraphics[width=0.10\linewidth]{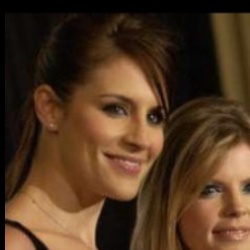}}
      \fbox{\includegraphics[width=0.10\linewidth]{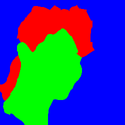}}
      \fbox{\includegraphics[width=0.10\linewidth]{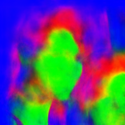}}
      \hfill
      \fbox{\includegraphics[width=0.10\linewidth]{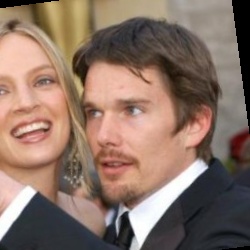}}
      \fbox{\includegraphics[width=0.10\linewidth]{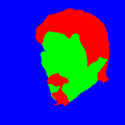}}
      \fbox{\includegraphics[width=0.10\linewidth]{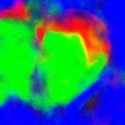}}\\
      \vspace{0.01\linewidth}
      \fbox{\includegraphics[width=0.10\linewidth]{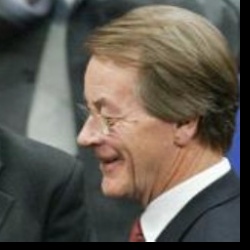}}
      \fbox{\includegraphics[width=0.10\linewidth]{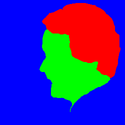}}
      \fbox{\includegraphics[width=0.10\linewidth]{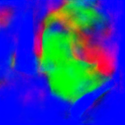}}
      \hfill
      \fbox{\includegraphics[width=0.10\linewidth]{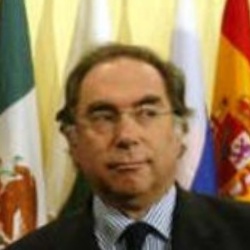}}
      \fbox{\includegraphics[width=0.10\linewidth]{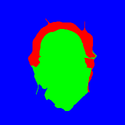}}
      \fbox{\includegraphics[width=0.10\linewidth]{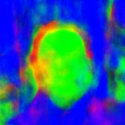}}
      \hfill
      \fbox{\includegraphics[width=0.10\linewidth]{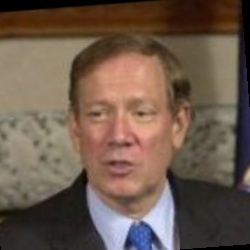}}
      \fbox{\includegraphics[width=0.10\linewidth]{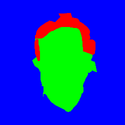}}
      \fbox{\includegraphics[width=0.10\linewidth]{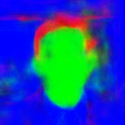}}\\
      \vspace{0.01\linewidth}
      \fbox{\includegraphics[width=0.10\linewidth]{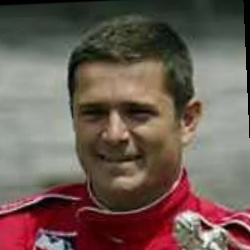}}
      \fbox{\includegraphics[width=0.10\linewidth]{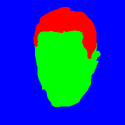}}
      \fbox{\includegraphics[width=0.10\linewidth]{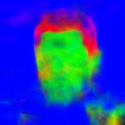}}
      \hfill
      \fbox{\includegraphics[width=0.10\linewidth]{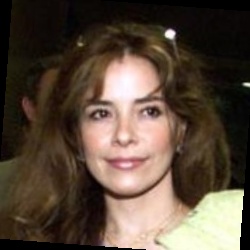}}
      \fbox{\includegraphics[width=0.10\linewidth]{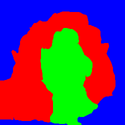}}
      \fbox{\includegraphics[width=0.10\linewidth]{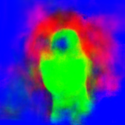}}
      \hfill
      \fbox{\includegraphics[width=0.10\linewidth]{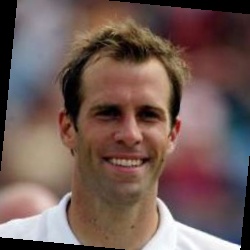}}
      \fbox{\includegraphics[width=0.10\linewidth]{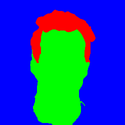}}
      \fbox{\includegraphics[width=0.10\linewidth]{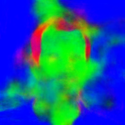}}\\
      \vspace{0.01\linewidth}
      \fbox{\includegraphics[width=0.10\linewidth]{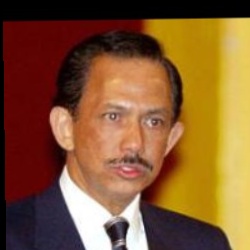}}
      \fbox{\includegraphics[width=0.10\linewidth]{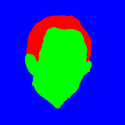}}
      \fbox{\includegraphics[width=0.10\linewidth]{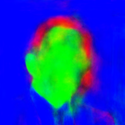}}
      \hfill
      \fbox{\includegraphics[width=0.10\linewidth]{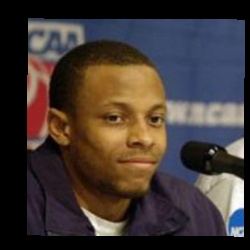}}
      \fbox{\includegraphics[width=0.10\linewidth]{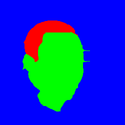}}
      \fbox{\includegraphics[width=0.10\linewidth]{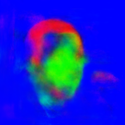}}
      \hfill
      \fbox{\includegraphics[width=0.10\linewidth]{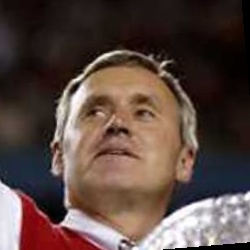}}
      \fbox{\includegraphics[width=0.10\linewidth]{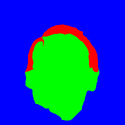}}
      \fbox{\includegraphics[width=0.10\linewidth]{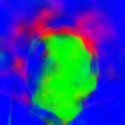}}\\
      \vspace{0.01\linewidth}
      \fbox{\includegraphics[width=0.10\linewidth]{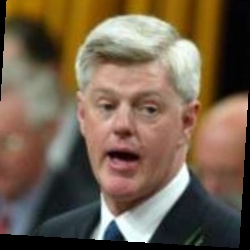}}
      \fbox{\includegraphics[width=0.10\linewidth]{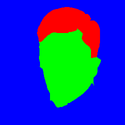}}
      \fbox{\includegraphics[width=0.10\linewidth]{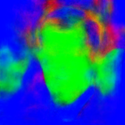}}
      \hfill
      \fbox{\includegraphics[width=0.10\linewidth]{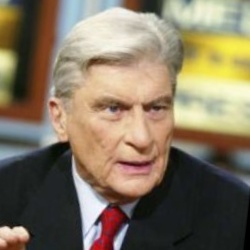}}
      \fbox{\includegraphics[width=0.10\linewidth]{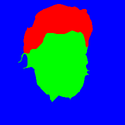}}
      \fbox{\includegraphics[width=0.10\linewidth]{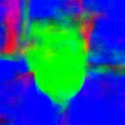}}
      \hfill
      \fbox{\includegraphics[width=0.10\linewidth]{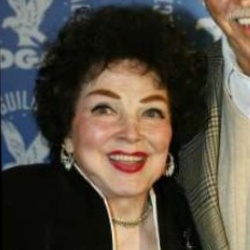}}
      \fbox{\includegraphics[width=0.10\linewidth]{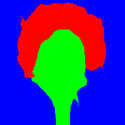}}
      \fbox{\includegraphics[width=0.10\linewidth]{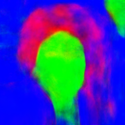}}\\
      \vspace{0.01\linewidth}
      \fbox{\includegraphics[width=0.10\linewidth]{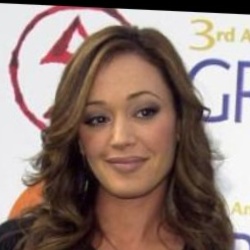}}
      \fbox{\includegraphics[width=0.10\linewidth]{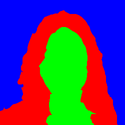}}
      \fbox{\includegraphics[width=0.10\linewidth]{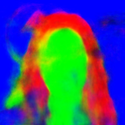}}
      \hfill
      \fbox{\includegraphics[width=0.10\linewidth]{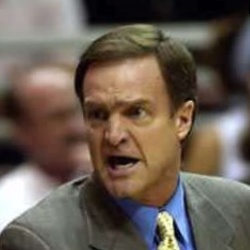}}
      \fbox{\includegraphics[width=0.10\linewidth]{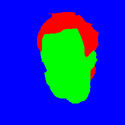}}
      \fbox{\includegraphics[width=0.10\linewidth]{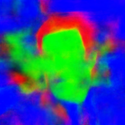}}
      \hfill
      \fbox{\includegraphics[width=0.10\linewidth]{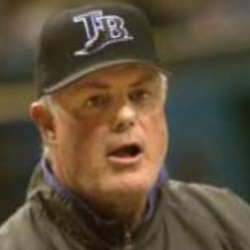}}
      \fbox{\includegraphics[width=0.10\linewidth]{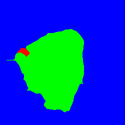}}
      \fbox{\includegraphics[width=0.10\linewidth]{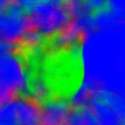}}\\
      \vspace{0.01\linewidth}
      \fbox{\includegraphics[width=0.10\linewidth]{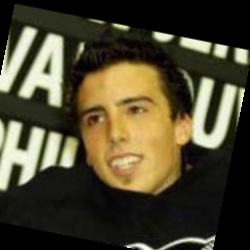}}
      \fbox{\includegraphics[width=0.10\linewidth]{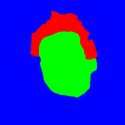}}
      \fbox{\includegraphics[width=0.10\linewidth]{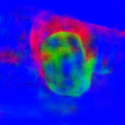}}
      \hfill
      \fbox{\includegraphics[width=0.10\linewidth]{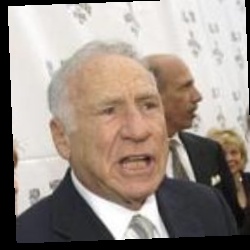}}
      \fbox{\includegraphics[width=0.10\linewidth]{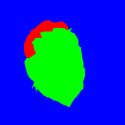}}
      \fbox{\includegraphics[width=0.10\linewidth]{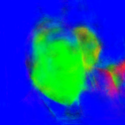}}
      \hfill
      \fbox{\includegraphics[width=0.10\linewidth]{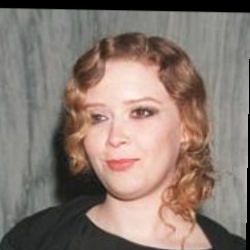}}
      \fbox{\includegraphics[width=0.10\linewidth]{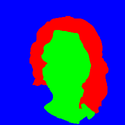}}
      \fbox{\includegraphics[width=0.10\linewidth]{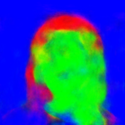}}
   \end{center}
   \vspace{-0.05\linewidth}
   \caption{
      \textbf{Part labeling results on LFW.}
      \emph{Left:}
         Image.
      \emph{Middle:}
         Ground-truth.  Semantic classes are skin (green), hair (red), and
         background (blue).
      \emph{Right:}
         Semantic labeling predicted via reconstructive sparse code transfer.
   }
   \label{fig:face_results}
\end{figure*}

Figure~\ref{fig:face_results} shows example results for semantic segmentation
of skin, hair, and background classes on the LFW parts dataset using
reconstructive sparse code transfer.  All results are for our two-layer
multipath network.  As the default split of the LFW parts dataset allowed
images of the same individual to appear in both training and test sets, we
randomly re-split the dataset with the constraint that images of a
particular individual were either all in the training set or all in the test
set, with no overlap.  All examples in Figure~\ref{fig:face_results} are from
our test set after this more stringent split.

Note that while faces are centered in the LFW part dataset images, we
directly apply our algorithm and make no attempt to take advantage of this
additional information.  Hence, for several examples in
Figure~\ref{fig:face_results} our learned skin and hair detectors fire on
both primary and secondary subjects appearing in the photograph.

\section{Conclusion}
\label{sec:conclusion}

We demonstrate that sparse coding, combined with a reconstructive transfer
learning framework, produces results competitive with the state-of-the-art
for contour detection.  Varying the target of the transfer learning stage
allows one to port a common sparse representation to multiple end tasks.  We
highlight semantic labeling of faces as an additional example.  Our approach
is entirely data-driven and relies on no hand-crafted features.  Sparse
representations similar to the one we consider also arise naturally
in the context of deep networks for image recognition.  We conjecture that
multipath sparse networks~\cite{MHMP} can produce shared representations
useful for many vision tasks and view this as a promising direction for future
research.

\vspace{0.020\textheight}

\noindent
\textbf{Acknowledgments.}
ARO/JPL-NASA Stennis NAS7.03001 supported Michael Maire's work.

\bibliographystyle{splncs}
\bibliography{sprec}

\begin{thebibliography}{10}

\bibitem{LRMDCCDN:ICML:2012}
Le, Q.V., Ranzato, M., Monga, R., Devin, M., Chen, K., Corrado, G.S., Dean, J.,
  Ng, A.Y.:
\newblock Building high-level features using large scale unsupervised learning.
\newblock ICML (2012)

\bibitem{LKF:ISCAS:2010}
LeCun, Y., Kavukcuoglu, K., Farabet, C.:
\newblock Convolutional networks and applications in vision.
\newblock ISCAS (2010)

\bibitem{KSH:NIPS:2012}
Krizhevsky, A., Sutskever, I., Hinton, G.E.:
\newblock {ImageNet} classification with deep convolutional neural networks.
\newblock NIPS (2012)

\bibitem{ZTF:ICCV:2011}
Zeiler, M.D., Taylor, G.W., Fergus, R.:
\newblock Adaptive deconvolutional networks for mid and high level feature
  learning.
\newblock ICCV (2011)

\bibitem{YLL:CVPR:2011}
Yu, K., Lin, Y., Lafferty, J.:
\newblock Learning image representations from the pixel level via hierarchical
  sparse coding.
\newblock CVPR (2011)

\bibitem{MHMP}
Bo, L., Ren, X., Fox, D.:
\newblock Multipath sparse coding using hierarchical matching pursuit.
\newblock CVPR (2013)

\bibitem{ZF:ECCV:2014}
Zeiler, M.D., Fergus, R.:
\newblock Visualizing and understanding convolutional networks.
\newblock ECCV (2014)

\bibitem{gPb-UCM}
Arbel\'{a}ez, P., Maire, M., Fowlkes, C., Malik, J.:
\newblock Contour detection and hierarchical image segmentation.
\newblock PAMI (2011)

\bibitem{EA:TIP:2006}
Elad, M., Aharon, M.:
\newblock Image denoising via sparse and redundant representations over learned
  dictionaries.
\newblock IEEE Transactions on Image Processing (2006)

\bibitem{OMP}
Pati, Y.C., Rezaiifar, R., Krishnaprasad, P.S.:
\newblock Orthogonal matching pursuit: Recursive function approximation with
  applications to wavelet decomposition.
\newblock Asilomar Conference on Signals, Systems and Computers (1993)

\bibitem{BOMP}
Rubinstein, R., Zibulevsky, M., Elad, M.:
\newblock Efficient implementation of the {K-SVD} algorithm using batch
  orthogonal matching pursuit.
\newblock (2008)

\bibitem{BSDS}
Martin, D., Fowlkes, C., Tal, D., Malik, J.:
\newblock A database of human segmented natural images and its application to
  evaluating segmentation algorithms and measuring ecological statistics.
\newblock ICCV (2001)

\bibitem{LFW}
Huang, G.B., Ramesh, M., Berg, T., Learned-Miller, E.:
\newblock Labeled faces in the wild: A database for studying face recognition
  in unconstrained environments.
\newblock (2007)

\bibitem{LFW-parts}
Kae, A., Sohn, K., Lee, H., Learned-Miller, E.:
\newblock Augmenting {CRF}s with {B}oltzmann machine shape priors for image
  labeling.
\newblock CVPR (2013)

\bibitem{SCG}
Ren, X., Bo, L.:
\newblock Discriminatively trained sparse code gradients for contour detection.
\newblock NIPS (2012)

\bibitem{SE}
Doll\'ar, P., Zitnick, C.L.:
\newblock Structured forests for fast edge detection.
\newblock ICCV (2013)

\bibitem{Pb}
Martin, D., Fowlkes, C., Malik, J.:
\newblock Learning to detect natural image boundaries using local brightness,
  color and texture cues.
\newblock PAMI (2004)

\bibitem{NCuts}
Shi, J., Malik, J.:
\newblock Normalized cuts and image segmentation.
\newblock PAMI (2000)

\bibitem{MBLS:IJCV:2001}
Malik, J., Belongie, S., Leung, T., Shi, J.:
\newblock Contour and texture analysis for image segmentation.
\newblock IJCV (2001)

\bibitem{RFM:ECCV:2006}
Ren, X., Fowlkes, C., Malik, J.:
\newblock Figure/ground assignment in natural images.
\newblock ECCV (2006)

\bibitem{LZD:CVPR:2013}
Lim, J., Zitnick, C.L., Doll\'ar, P.:
\newblock Sketch tokens: A learned mid-level representation for contour and
  object detection.
\newblock CVPR (2013)

\bibitem{MLBHP:ECCV:2008}
Mairal, J., Leordeanu, M., Bach, F., Hebert, M., Ponce, J.:
\newblock Discriminative sparse image models for class-specific edge detection
  and image interpretation.
\newblock ECCV (2008)

\bibitem{YWLSH:CVPR:2012}
Yang, J., Wang, Z., Lin, Z., Shu, X., Huang, T.:
\newblock Bilevel sparse coding for coupled feature spaces.
\newblock CVPR (2012)

\bibitem{KSVD}
Aharon, M., Elad, M., Bruckstein, A.:
\newblock {K-SVD}: An algorithm for designing overcomplete dictionaries for
  sparse representation.
\newblock IEEE Transactions on Signal Processing (2006)

\bibitem{MYP:ICCV:2013}
Maire, M., Yu, S.X., Perona, P.:
\newblock Progressive multigrid eigensolvers for multiscale spectral
  segmentation.
\newblock ICCV (2013)

\bibitem{UCM}
Arbel\'{a}ez, P.:
\newblock Boundary extraction in natural images using ultrametric contour maps.
\newblock POCV (2006)

\bibitem{Canny}
Canny, J.:
\newblock A computational approach to edge detection.
\newblock PAMI (1986)

\end{thebibliography}

\end{document}